\documentclass[journal]{IEEEtran}
\usepackage{cite}
\usepackage{graphicx}
\usepackage{url}
\usepackage{amsmath}
\usepackage[caption = false, font=footnotesize]{subfig}
\usepackage{float}
\usepackage{booktabs} % Allows the use of \toprule, \midrule and \bottomrule in tables for horizontal lines
\usepackage{multirow}
\usepackage{array}
\usepackage{color}
\newcolumntype{L}[1]{>{\raggedright\let\newline\\\arraybackslash\hspace{0pt}}m{#1}}
\newcolumntype{C}[1]{>{\centering\let\newline\\\arraybackslash\hspace{0pt}}m{#1}}
\newcolumntype{R}[1]{>{\raggedleft\let\newline\\\arraybackslash\hspace{0pt}}m{#1}}

\newcommand{\ie}{\textit{i}.\textit{e}.}
\newcommand{\eg}{\textit{e}.\textit{g}.}

\ifCLASSINFOpdf
\else
\fi
\begin{document}
\title{Deep Blur Mapping: Exploiting \\ High-Level Semantics by  Deep Neural Networks}

\author{Kede~Ma,~\IEEEmembership{Member,~IEEE,}
        Huan~Fu,
        Tongliang~Liu,~\IEEEmembership{Member,~IEEE,}\\
        Zhou~Wang,~\IEEEmembership{Fellow,~IEEE,}
        and~Dacheng~Tao,~\IEEEmembership{Fellow,~IEEE}% <-this % stops a space
\thanks{This work was supported in part by the Natural
Sciences and Engineering Research Council of Canada, and in part by  the Australian Research Council Projects FL-170100117, DP-180103424, and LP-150100671.}
\thanks{Kede Ma and Zhou Wang are with the Department of Electrical and Computer Engineering, University of Waterloo, Waterloo, ON N2L 3G1, Canada (e-mail: k29ma@uwaterloo.ca; zhou.wang@uwaterloo.ca).}
\thanks{Huan Fu, Tongliang Liu, and Dacheng Tao are with the UBTECH Sydney Artificial Intelligence Centre and the School of Information Technologies, the Faculty of Engineering and Information Technologies, The University of Sydney, Sydney, NSW 2008, Australia (e-mail: hufu6371@uni.sydney.edu.au; tongliang.liu@sydney.edu.au; dacheng.tao@sydney.edu.au).}% <-this % stops a space
%\thanks{}% <-this % stops a space
%\thanks{}
}


\markboth{}%
{Shell \MakeLowercase{\textit{et al.}}: Bare Demo of IEEEtran.cls for Journals}



\maketitle

\begin{abstract}
  The human visual system excels at detecting local blur of visual images, but the underlying mechanism is not well understood. Traditional views of blur such as reduction in energy at high frequencies and loss of phase coherence at  localized features have fundamental limitations. For example, they cannot well discriminate flat regions from blurred ones. Here we propose that high-level semantic information is critical in successfully identifying local blur. Therefore, we resort to deep neural networks that are proficient at learning high-level features and propose the first end-to-end local blur mapping algorithm based on a fully convolutional network. By analyzing various architectures with different depths and design philosophies,  we empirically show that high-level features of deeper layers play a more important role than low-level features of shallower layers in resolving challenging ambiguities for this task.  We test the proposed method on a standard blur detection benchmark and demonstrate that it significantly advances the state-of-the-art (ODS F-score of $0.853$). Furthermore, we explore the use of the generated blur maps in three applications, including blur region segmentation, blur degree estimation, and blur magnification.
  \end{abstract}


% \fi
\IEEEpeerreviewmaketitle

%%%%%%%%% BODY TEXT
\section{Introduction}
\IEEEPARstart{B}{lur} is one of the most common image degradations that arises from a number of sources, including atmospheric scatter, camera shake, defocus, and object motion. It is also manipulated by photographers to create visually pleasing effect that draws attention to humans/objects of interest. Given a natural photographic image, the goal of local blur mapping is to label every pixel as either blurry or non-blurry, resulting in a blur map. Local blur mapping is an important component in many image processing and computer vision systems. For image quality assessment, blur is an indispensable factor that affects perceptual image quality~\cite{wang2004image,ma2016group}. For example, the worst quality images scored by human subjects in the LIVE Challenge Database~\cite{ghadiyaram2016massive} mainly suffer from motion and/or out-of-focus blur. For object detection, the identified blurred regions may be excluded for efficient and robust object localization~\cite{ren2015faster}. Other applications that may benefit from local blur mapping include image restoration~\cite{dai2009removing,efrat2013accurate}, photo editing~\cite{bae2007defocus}, depth recovery~\cite{mather1997use,shi2015just}, and image segmentation~\cite{favaro2004variational}.
\begin{figure}
  \centering
  \includegraphics[width=.5\textwidth]{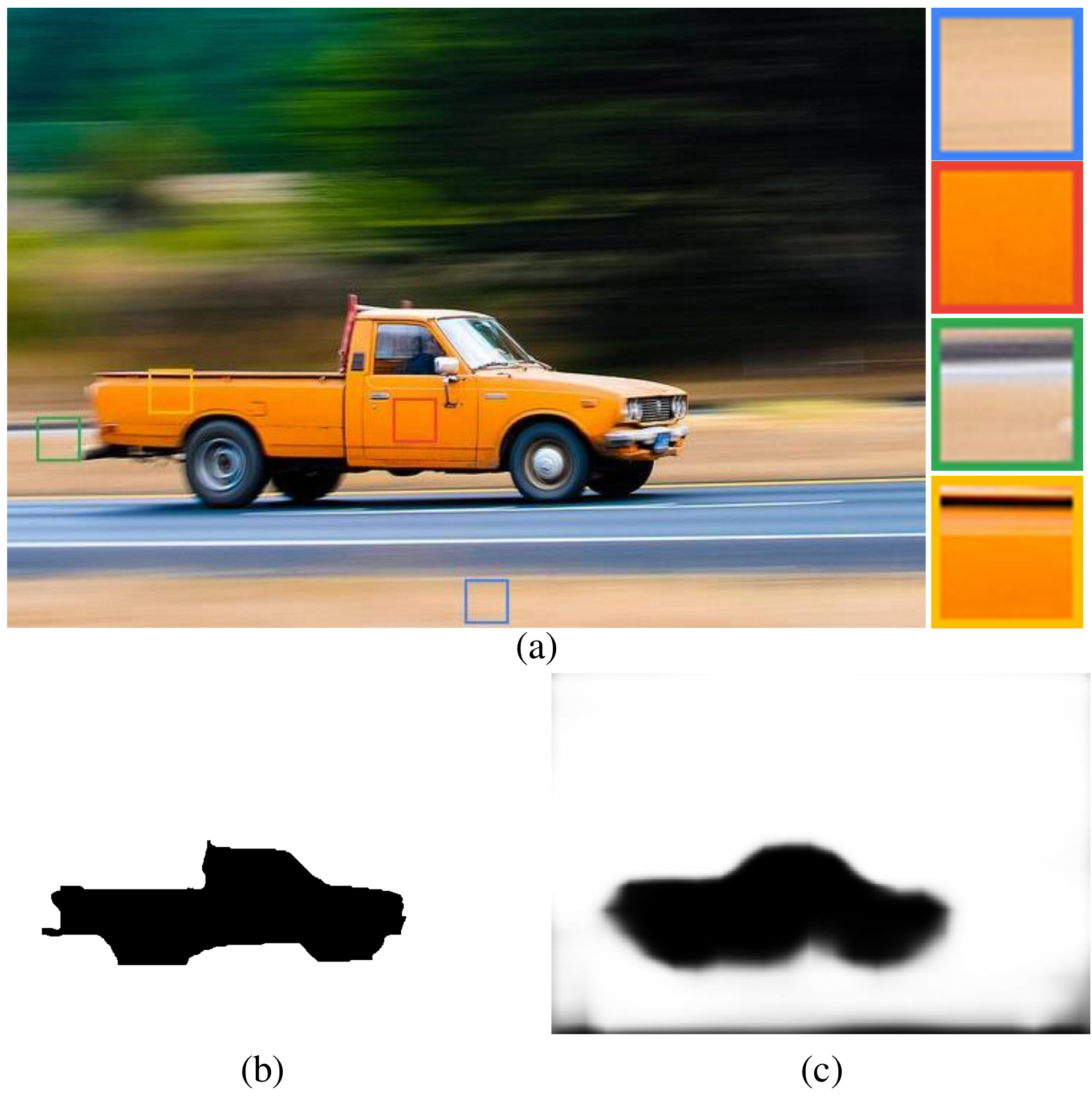}
  \caption{Challenges in local blur mapping. The pairs of (blue, red) and (green, yellow) framed patches appear to be similar in terms of local structural features and complexities, making them difficult for local feature-based approaches to identify blur. By contrast, semantic information is helpful in making the distinction. (a) Test image from the blur detection benchmark~\cite{shi2014discriminative}. (b) Ground truth blur map. (c) Blur map produced by the proposed DBM.}\label{fig:ambiguity}
\end{figure}
%\begin{figure}
%    \centering
%    \captionsetup{justification=centering}
%
%
%    \subfloat[A motion blurred image]
%    {\includegraphics[width=0.48\textwidth]{figs/output_motion0044}}\hskip.2em
%    %\subfloat[]
%    % {\includegraphics[width=0.2\textwidth]{figs/motion0044}}\hskip.2em
%    \subfloat[Blurred]{\includegraphics[width=0.11\textwidth]{figs/1}}\hskip.5em
%    \subfloat[Flat]{\includegraphics[width=0.11\textwidth]{figs/2}}\hskip.5em
%    \subfloat[Edge]{\includegraphics[width=0.11\textwidth]{figs/4}}\hskip.5em
%    \subfloat[Residual]{\includegraphics[width=0.11\textwidth]{figs/3}}
%
%    \caption{Illustration of the challenging ambiguities in image partial blur detection. (a) shows an example test image in the blur detection benchmark~\cite{shi2014discriminative}; (b) and (c) show the similar appearance between the blurred and flat regions; (d) and (e) show the indistinguishable structures between sharp edges without blurring and edge residuals after blurring. }\label{fig:ambiguity}
%\end{figure}

The human visual system (HVS) is good at identifying the blurry parts of an image with amazing speed~\cite{webster2002neural,wandell1995foundations}, but the underlying mechanism is not well understood.
A traditional view of blur is that it reduces the energy (either globally or locally) at high frequencies. Several low-level features have been  hand-crafted to exploit this observation. Among those, power spectral slopes~\cite{liu2008image,shi2014discriminative} and image gradient statistics~\cite{levin2006blind,liu2008image,shi2014discriminative} are representative. Another view of blur perception is that it arises from the disruption of the local phase coherence at precisely localized features (\eg, step edges)~\cite{wang2003local}. Therefore, a coarse-to-fine phase prediction may serve as an indication of blur~\cite{wang2003local}. Nearly all previous local blur mappers
%~\cite{liu2008image,chakrabarti2010analyzing,zhuo2011defocus,su2011blurred,shi2014discriminative,zhang2015fast,pang2016classifying,zhang2016spatially,chen2016fast,tang2016spectral,yi2016lbp,zhu2016efficient,alireza2017spatially,javaran2017automatic} 
rely on the two assumptions either explicitly or implicitly with limited success. In particular, they fail to  discriminate flat and blurred regions, and  they often mix up structures with and without blurring. A visual example is shown in Fig.~\ref{fig:ambiguity}, where we can see that both the blue and red framed patches appear to be smooth but from different origins. Specifically, the textures of sand in the blue framed patch are lost due to the severe blur, while the car body in the red framed patch is flat in nature. On the other hand, the green and yellow framed patches have similar local structural features at the top with similar complexities, but the former suffers from blur, while the latter does not.  All of these make the local blur mapping task difficult for  local feature-based algorithms.

%In this work, we provide an alternative explanation for blur perception: blur arises when we have difficulty parsing the images of natural scenes we encounter in our daily life. On the one hand, if we cannot parse some parts of the image, those regions are probably blurred.  On the other hand, even though we may be able to infer the underlying structures based on the context, they do not appear in ways they should be in the real world, and we also perceive them as blurry. In either case,

In this regard, we argue that the fundamental problem in existing approaches is their ignorance to high-level semantic information in natural images, which is crucial in successfully identifying local blur. Therefore, we resort to deep convolutional neural networks (CNN) that have advanced the state-of-the-art in many high-level vision tasks such as image classification~\cite{He2015}, object detection~\cite{ren2015faster}, and semantic segmentation~\cite{long2015fully}. Specifically, we develop the first fully convolutional network (FCN)~\cite{long2015fully} for end-to-end and image-to-image blur mapping~\cite{shi2014discriminative}, which we name deep blur mapper (DBM). By fully convolutional, we mean all the learnable filters in the network are convolutional and no fully connected layers are involved. As a result, DBM allows input of arbitrary size, encodes spatial information thoroughly for better prediction, and maintains a relatively low computational cost. We adopt various architectures with different depths by trimming the 16-layer VGGNet~\cite{Simonyan2015Very} from different convolutional stages. By doing so, we empirically show that high-level features from deeper layers are more important than low-level features from shallower layers in resolving challenging ambiguities for local blur mapping, which conforms to our perspective of blur perception. We also experiment with more advanced directed acyclic graph based architectures that better combine low-level spatial and high-level semantic information but yield no substantial improvement, which again verify the critical role of high-level semantics in this task. Due to the limited number of training samples, we initialize all networks with weights pre-trained on the semantic segmentation task~\cite{long2015fully} that contain rich high-level information about what an input image constitutes. DBM is tested on a standard blur detection benchmark~\cite{shi2014discriminative} and outperforms state-of-the-art methods by a large margin.

Our contributions are three-fold. First, we provide a new perspective on blur perception, where high-level semantic information plays a critical role. Second, we show that it is possible to learn an end-to-end and image-to-image local blur mapper based on FCNs~\cite{long2015fully}, which well addresses challenging ambiguities such as differentiating flat and blurred regions, and structures with and without blurring. Third, we explore three potential applications of the generated blur maps, \ie, blur region segmentation, blur degree estimation,  and blur magnification.

The rest of the paper is organized in the following manner. Section~\ref{sec:rw} reviews the related work of local blur mapping with  emphasis on statistical analysis of traditional hand-crafted low-level features. Section~\ref{sec:method} details the proposed DBM based on FCNs and their alternative architectures.  Section~\ref{sec:experiment} conducts extensive comparative and ablation experiments to validate the promise of DBM. Section~\ref{sec:conclusion} concludes the paper.

\section{Related Work}\label{sec:rw}
The computational blur analysis is a long-standing problem in vision and image processing research, and early works can be dated back to as early as 1960s~\cite{slepian1967restoration}. Most researchers in this field focus on the image deblurring problem that aims to restore a sharp image from a blurred version~\cite{cannon1976blind,xu2010two}. As an ill-posed problem, many algorithms assume uniform camera motion and the availability of the structure of the point spread function (PSF). Blind image deblurring takes a step further by simultaneously recovering the PSF and the latent unblurred image. It is frequently cast as a maximum a posteriori estimation problem, characterizing the unblurred image using natural image statistics as priors~\cite{miskin2000ensemble,levin2006blind,levin2007image}. In practice, the PSF is often spatially varying, making blind image deblurring algorithms unstable and unsatisfactory. On the contrary, blur mapping itself is little investigated. Early works on blur mapping quantify the overall blur degree of an image and cannot perform dense prediction. For example, Marziliano {\textit{et al}.}  analyzed the spread of the edges~\cite{marziliano2002no}. A similar approach was proposed in~\cite{elder1998local} by estimating the thickness of object contours. Zhang and Bergholm~\cite{zhang1997multi} designed a Gaussian difference signature to model the diffuseness caused by out-of-focus blur. The images under consideration are usually uniformly blurred with a Gaussian PSF, and therefore the results can be directly linked to perceptual quality, but cannot be generalized to non-Gaussian and non-uniform blur cases in the real world. 

\begin{figure}[t]
    \centering
    \captionsetup{justification=centering}
    \subfloat[]
    {\includegraphics[width=0.5\textwidth]{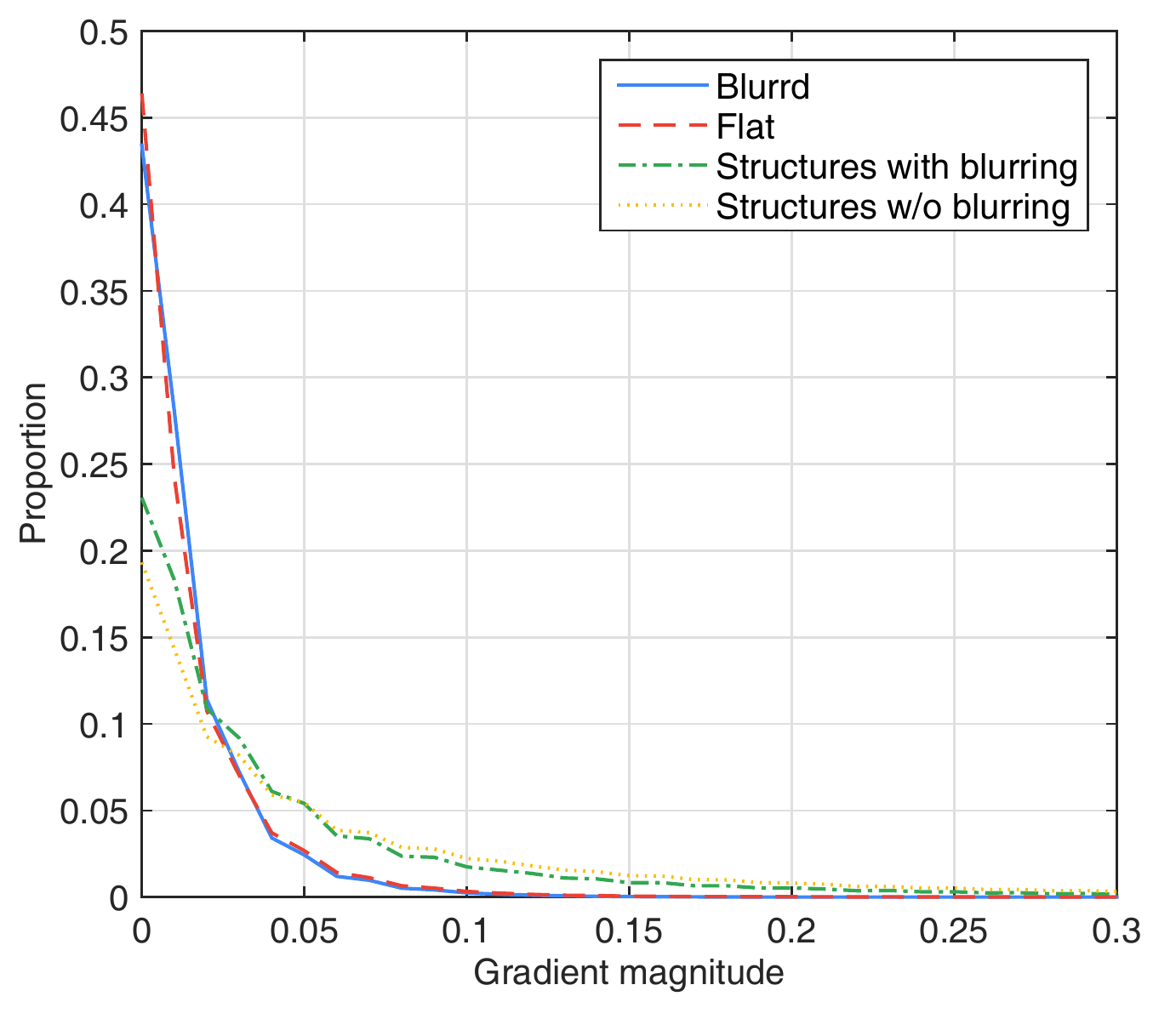}}\hskip.2em
     \subfloat[]{\includegraphics[width=0.5\textwidth]{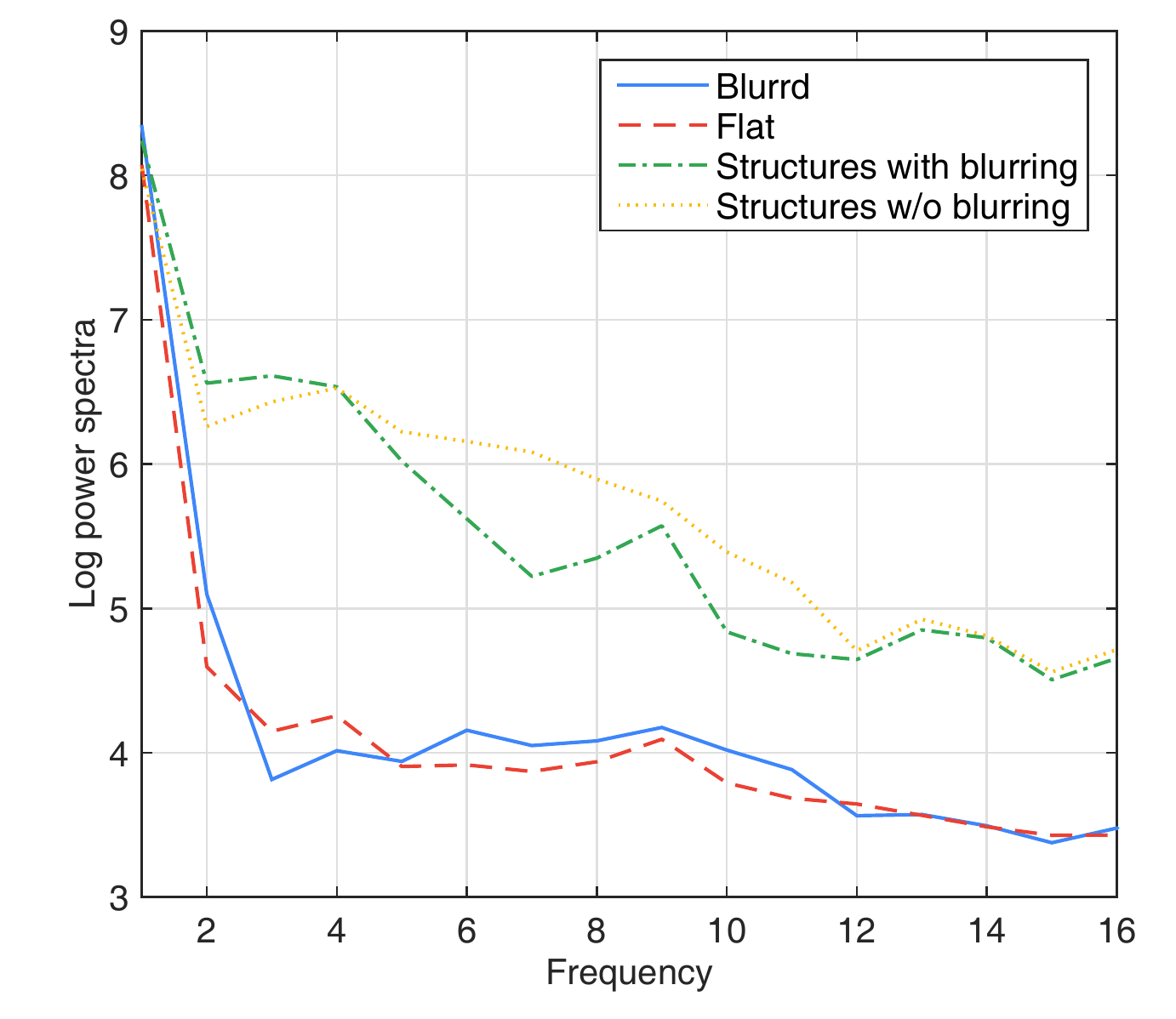}}
    \caption{Traditional low-level features fail to differentiate between flat and blurry regions, and structures with and without blurring. (a) Local gradient statistics. (b) Local power spectral slopes. Although the gradient distribution of blurred patches exhibits a sharp peak at zero and a less heavy tail, which is distinctive from structured patches, it cannot tell for example whether structured patches undergo blurring or not. Similar is observed for local power spectral slopes. We extract patches from $100$ images in the blur detection benchmark~\cite{shi2014discriminative} to draw (a) and use the four patches in Fig.~\ref{fig:ambiguity} to draw (b).   }\label{fig:blurSta}
\end{figure}

Only recently has local blur mapping become an active research
topic. Rugna and Konik~\cite{da2003automatic} identified blurry regions by exploiting the observation that they are more invariant to low-pass filtering. Blind deconvolution-based methods have also been investigated to segment motion-blurred~\cite{levin2006blind} and defocus-blurred~\cite{kovacs2007focus} regions. Zhuo and Sim~\cite{zhuo2011defocus} exploited the fact that the difference between the blurred and re-blurred patches is insignificant, and estimated the defocus blur amount by the ratio between the gradients of input and re-blurred images. 
Javaran \textit{et al.}~\cite{javaran2017automatic} practiced similar ideas in~\cite{zhuo2011defocus} and characterized the difference in the DCT domain. Su {\textit{et al}.}~\cite{su2011blurred} examined the singular value information between blurry and non-blurry regions. Chakrabarti {\textit{et al}.}~\cite{chakrabarti2010analyzing} adopted local Fourier transform to analyze directional blur. Liu {\textit{et al}.}~\cite{liu2008image} manually designed three local features represented by spectrum, gradient, and color information for blurry region extraction. Their features have been later improved by Shi {\textit{et al}.}~\cite{shi2014discriminative} and combined with responses of learned local filters to jointly analyze blurry regions in a multi-scale fashion.  
%Based on  double discrete wavelet transform, Zhang and Hirakawa~\cite{zhang2015fast} recovered a dense estimation of linear blur kernel in the luminance domain. 
Pang \textit{et al.}~\cite{pang2016classifying} described spatially varying blur by kernel-specific features. 
% Zhang \textit{et al.}~\cite{zhang2016spatially} estimated the defocus blur map based on edge information follow by matting interpolation. 
Chen~\cite{chen2016fast} \textit{et al.} adopted fast defocus belief propagation. Tang \textit{et al.}~\cite{tang2016spectral} presented a blur metric
based on the log averaged spectrum residual, whose maps are  coarse-to-fine refined  by exploiting  neighborhood similarity.
 Yi and Eramian \textit{et al.}~\cite{yi2016lbp} explored local binary patterns in the context of blur identification and found that  blurry regions have fewer local binary patterns compared with those of sharp regions. Zhu and Karam~\cite{zhu2016efficient} quantified the level of spatially varying blur by integrating directional edge spread and just noticeable blur. More recently, Golestaneh and Karam~\cite{alireza2017spatially} computed blur detection maps based on a high-frequency multi-scale fusion and sort transform of gradient magnitudes.

All the above-mentioned methods are based on hand-crafted low-level features and cannot robustly tell which parts of an image are truly blurred or flat in nature, and which parts of structures have been blurred or not. To have a closer look, we perform statistical analysis of two representative low-level features, namely local gradient statistics and local power spectral slopes, as shown in Fig.~\ref{fig:blurSta}. The local gradient statistics are drawn by extracting more than $1$ million patches from $100$ images in the blur detection benchmark~\cite{shi2014discriminative}. It is widely recognized that the gradient distribution of blurred patches exhibits a sharper peak at zero and a less heavier tail than those of structured patches~\cite{simoncelli2001natural}. However, it cannot tell for example whether structured patches undergo blurring or not, and whether smooth patches are flat in nature or severely blurred.  Similar is true for local power spectral slope based measures, for example, on the four patches in Fig.~\ref{fig:ambiguity}.

A closely related area to blur mapping is image sharpness measurement~\cite{ferzli2009no,hassen2013image}, which targets at extracting sharp regions from an image. The results may be combined to an overall sharpness score (global assessment) or refined to a sharpness map (local assessment). Most existing sharpness measurement algorithms rely on similar assumptions to those of blur mapping, but there are subtle differences. For example, in sharpness assessment, flat and blurry regions can both be regarded as non-sharp, but in blur mapping, discriminating them is a must for a successful algorithm.

\section{Deep Blur Mapper}\label{sec:method}
%The proposed FCN-based method that exploits high-level features is an instantiation of our new view of blur perception that blur arises when the HVS finds difficulty parsing the image. Under this assumption, other algorithms that makes proper use of high-level information when performing local blur mapping should also work.

At a high level, we feed an image of arbitrary size into an FCN and the network successively outputs a blur map of the same size with each entry ranging between $[0,1]$ to indicate the probability of the corresponding pixel being blurred. The size mismatch is resolved by in-network upsampling. Through a standard stochastic gradient descent (SGD) training procedure, our network is able to learn a complex mapping from raw image pixels to blur perception.
\subsection{Training and Testing}
Given a training image set $\mathcal{B}=\{({\bf X}_k, {\bf Y}_k)\}_{k = 1}^{K}$, where ${\bf X}_k$ is the $k$-th raw input image and ${\bf Y}_k$ is the corresponding ground truth binary blur map, our goal is to learn an FCN ${\hat{\bf Y}_k} = f({\bf X}_k)$ that produces a blur map with high accuracy. It is convenient to drop the subscript $k$ without ambiguity due to the image-wise operation. We denote all layer parameters in the network as ${\bf W}$. The loss function is a sum over per-pixel losses between the prediction ${\hat{\bf Y}} = \{{\hat y}_i, i = 1,\cdots, |Y|\}$ and the ground truth ${\bf Y }= \{y_i \in \{0,1\}, i = 1,\cdots, |Y|\}$, where $i$ indicates the spatial coordinate. We consider the cross entropy loss
\begin{equation}\label{eq:lso}
\begin{split}
\ell({\bf W}) =& -\sum_{i = 1}^{|Y|}y_i\log\Pr ({\hat y}_i = 1|{\bf X}, {\bf W})\\
&-(1-y_i)\log\Pr ({\hat y}_i = 0|{\bf X}, {\bf W})\,.
\end{split}
\end{equation}
$\Pr ({\hat y}_i = 1|{\bf X}, {\bf W}) \in [0,1]$ is implemented by the sigmoid function  on the $i$-th activation.
Eq.~(\ref{eq:lso}) can be easily extended to account for the class imbalance situation by weighting the loss according to the proportion of positive and negative labels. Although the labels in the blur detection database~\cite{shi2014discriminative} are mildly unbalanced (around $64\%$ pixels are blurred), we find using the class-balanced cross entropy loss unnecessary. In addition, many probability distribution
measures can be adopted as alternatives to the cross entropy
loss, such as the fidelity loss from quantum physics~\cite{nielsen2010quantum}. We find in our experiments
that the fidelity loss gives very similar performance. Therefore, we choose the cross entropy loss throughout the paper.

After training, the optimal layer parameters ${{\bf W}^{\star}}$ are learned. Given a test image ${\bf X}$, we perform a standard forward pass to obtain the predicted blur map:
\begin{equation}\label{eq:fwd}
{\hat {\bf Y} } = f({\bf X},{{\bf W}^\star}  )\,.
\end{equation}

\begin{table}
  \centering
  \caption{The FCN configurations. The depth increases from left to right.  We follow the convention in~\cite{Simonyan2015Very} and denote the weights of a convolutional layer as ``conv$\langle$receptive field size$\rangle$-$\langle$number of channels$\rangle$''. The ReLU nonlinearity is omitted here for brevity}\label{tab:configuration}
  \begin{tabular}{c|c|c|c|c}
      \toprule
    \multicolumn{5}{c}{FCN configuration}\\
     \hline
      I&II&III&IV&V\\
      \hline
      3 weight&5 weight&8 weight&11 weight&14 weight\\
      layers&layers&layers&layers&layers\\
      \hline
      \multicolumn{5}{c}{input image of arbitrary size}\\
      \hline
      conv3-64&conv3-64&conv3-64&conv3-64&conv3-64\\
      conv3-64&conv3-64&conv3-64&conv3-64&conv3-64\\
      \cline{2-5}
      conv1-1& \multicolumn{4}{c}{max-pooling}\\
      \hline
      &conv3-128&conv3-128&conv3-128&conv3-128\\
    
      &conv3-128&conv3-128&conv3-128&conv3-128\\
      \cline{3-5}
      &conv1-1&\multicolumn{3}{c}{max-pooling}\\
       \hline
       & &conv3-256&conv3-256&conv3-256\\
       & &conv3-256&conv3-256&conv3-256\\
       & &conv3-256&conv3-256&conv3-256\\
      \cline{4-5}
      & &conv1-1&\multicolumn{2}{c}{max-pooling}\\
      \hline
      &&&conv3-512&conv3-512\\
      &&&conv3-512&conv3-512\\
      &&&conv3-512&conv3-512\\
      \cline{5-5}
      & & &conv1-1&\multicolumn{1}{c}{max-pooling}\\
       \hline
       & &&&conv3-512\\
       & &&&conv3-512\\
       & &&&conv3-512\\
       & &&&conv1-1\\
       \hline
       \multicolumn{5}{c}{in-network upsampling}\\
       \hline
       \multicolumn{5}{c}{sigmoid}\\
     \bottomrule
   \end{tabular}
\end{table}
\subsection{Network Architectures and Alternatives}
\begin{figure}
  \centering
  \includegraphics[width=0.5\textwidth]{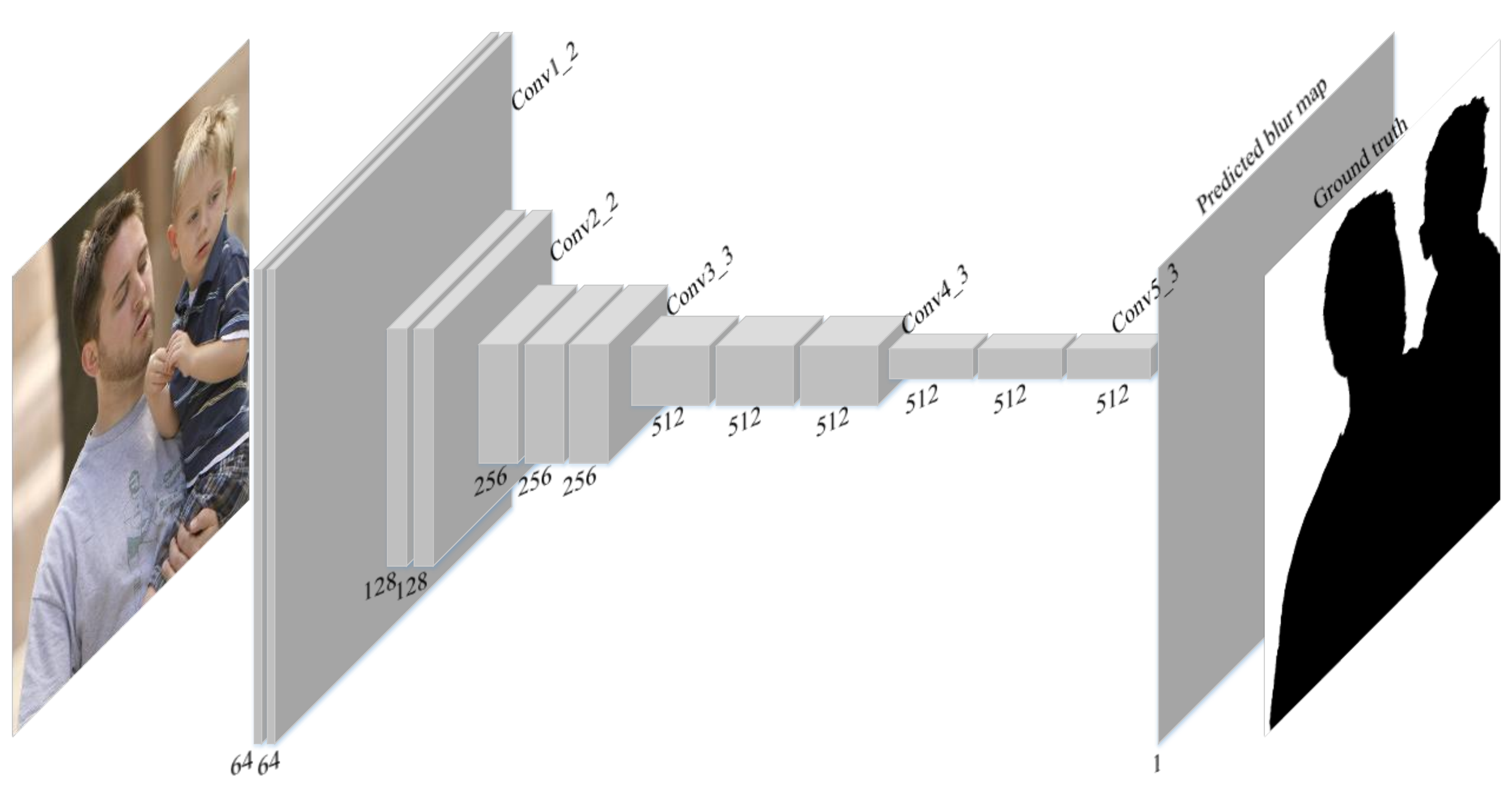}
  \caption{Configuration V as the default architecture for local blur mapping by trimming the VGG16 network. The height and width of the boxes represent the spatial sizes of the filter responses, which depend upon the size of the input image.  The depth of the boxes indicates the filter number used in each layer. Here, we omit ReLU and max-pooling layers for simplicity. After the last convolutional layer ($\text{conv5}\_\text{3}$), our mapper performs a $16\times$ in-network upsampling to obtain the final blur map.}\label{fig:architecture}
\end{figure}

Inspired by recent works~\cite{bertasius2015deepedge,xie2015holistically} that successfully fine-tune deep neural networks pre-trained on  image classification for edge detection, we analyze several linearly cascaded FCNs based on the 16-layer VGGNet architecture~\cite{Simonyan2015Very}, which has been extremely popular and extensively studied in image classification. Specifically, we trim the VGGNet up to the last convolutional layer in each stage, \ie, $\text{conv1}\_\text{2}$, $\text{conv2}\_\text{2}$, $\text{conv3}\_\text{3}$, $\text{conv4}\_\text{3}$, and $\text{conv5}\_\text{3}$, respectively, resulting in five FCNs with different depths. For each network, we add a convolutional layer with a $1\times 1$ receptive field,  which performs a linear summation of the input channels. Finally, we implement an in-network upsampling followed by sigmoid nonlinearity  to counteract the size mismatch between the generated and ground truth blur maps.  For each configuration, we throw away all fully connected layers to make it fully convolutional as in~\cite{long2015fully}, because it significantly reduces the computational complexity with only mild loss of representation power. As a result, we speed up the computation and reduce the memory storage at both the training and test stages. Moreover, we choose to drop the max-pooling layer immediately after the last convolutional layer at each stage, trying to keep as finer spatial information as possible to make later interpolation easier. The detailed configurations of the five FCNs are summarized in Table~\ref{tab:configuration} and Configuration V that is used as the default architecture for performance comparison is shown in Fig.~\ref{fig:architecture}. 

The five network configurations characterized by different depths favor different types of information. Configuration I retains the spatial information intact, which is ideal for dense prediction. However, due to its shallow structure, it can only extract low-level information and fail to learn powerful semantic information from the image. On the contrary, Configuration V has a very deep structure, which consists of  $14$ stacks of convolutional filters. Therefore, it is capable of transforming low-level features into high-level semantic information, but sacrifices  fine spatial information due to max-pooling. The in-network upsampling has to be performed in order to recover the spatial resolution. With the five configurations, we are able to empirically study the relative importance of spatial and semantic information in local blur mapping. As will be clear in Section~\ref{subsubsec:config}, semantic information plays a dominant role in local blur mapping. By contrast, spatial information is less relevant.

We continue by discussing several more sophisticated architecture designs that better combine low-level and high-level features. We first briefly introduce FCNs with skip layers. The original FCNs make use of classification nets for dense semantic segmentation~\cite{long2015fully} by transferring fully connected layers into convolutional ones. To combat the coarse spatial information in deeper layers, which limits the scale of the details in the upsampled output, Long {\textit{et al}.}~\cite{long2015fully} introduced skip layers that combine the responses of the final prediction layer with those of shallower layers with finer spatial information. It is straightforward to adapt this architecture to the blur mapping task and we include FCN-8s, a top-performing architecture with reasonable complexity in our experiment. Moreover, to make the learning process of hidden layers direct and transparent, Lee {\textit{et al}.}~\cite{lee2015deeply} proposed deeply supervised nets (DSN) that add side output layers to the convolutional layers in each stage. In the case of 16-layer VGGNet adopted in edge detection~\cite{xie2015holistically}, five side outputs are produced right after $\text{conv1}\_\text{2}$, $\text{conv2}\_\text{2}$, $\text{conv3}\_\text{3}$, $\text{conv4}\_\text{3}$, and $\text{conv5}\_\text{3}$ layers, respectively. All side outputs are fused to a final output, whose weights are learnable. The final output together with all side outputs contribute to the loss. We include two variants of DSN: training with weighted fusion only, and training with weighted fusion and deep supervision.  As will be clear in Section~\ref{subsubsec:high}, incorporating low-level features through these sophisticated  architectures often impairs performance compared with the default Configuration V.

\section{Experiments}\label{sec:experiment}
In this section, we first provide thorough implementation details on training and testing the proposed DBM. We then describe the experimental protocol and analyze the five FCN configurations, from which we choose Configuration V as our default architecture to compare with nine state-of-the-art methods. Finally, we analyze various aspects of DBM with  emphasis on the role of high-level semantics. All models are trained and tested with Caffe~\cite{jia2014caffe}. 

\subsection{Implementations}
We first describe data preparation and preprocessing. To the best of our knowledge,  the blur detection benchmark built by Shi {\textit{et al}.}~\cite{shi2014discriminative} is the only  database that is publicly available for this task. It
contains $1,000$ images with human labeled blur regions, among which $296$ are partially motion-blurred and $704$ are defocus-blurred. Since the number of training samples is limited, we only divide it into training and test sets, without using the validation set for hyper-parameter tuning. It turns out that the only critical hyper-parameter is the learning rate and as long as we set it to a reasonably small value that keeps the gradients from blowing up, no noticeable differences in the final results are observed. The training set contains images with odd indices, denoted by $D_o$ and the test set contains images with even indices, denoted by $D_e$. DBM allows for input of arbitrary size, so we try various input sizes and find that it is insensitive to input size variations. We take advantage of this observation and resize all images to $384\times 384$ in order to reduce GPU memory cost and speed up training and testing. Another preprocessing step is to subtract the mean RGB value, computed on the ImageNet database~\cite{deng2009imagenet}. We also try to augment training samples by incorporating modest geometric (flipping, rotating, and scaling) and photometric (brightness, contrast, saturation, and gamma mapping) transformations without hurting their perceptual quality and high-level semantics, but this does not yield noticeable improvement. Therefore, the reported results in the paper are without data augmentation.

We initialize the layer parameters with weights from a full 16-layer VGGNet pre-trained on the semantic segmentation task~\cite{long2015fully} and fine-tune them by SGD with momentum. The training is regularized by weight decay (the $l_2$ penalty multiplier set to $5 \times 10^{-4}$). The learning rate is initially set to be $2^{-10}$ and follows a polynomial decay with a power of $0.9$. The learning rate for biases is doubled. The batch size is set to $3$ images, and momentum to $0.9$. The in-network upsampling layer is fixed to bilinear interpolation. Although the interpolation weights are learnable, the additional performance gain is marginal.  The learning stops when the maximum iteration number $10,000$ is reached. The final weights are used for testing.

 \begin{figure}
  \centering
  \includegraphics[width=.5\textwidth]{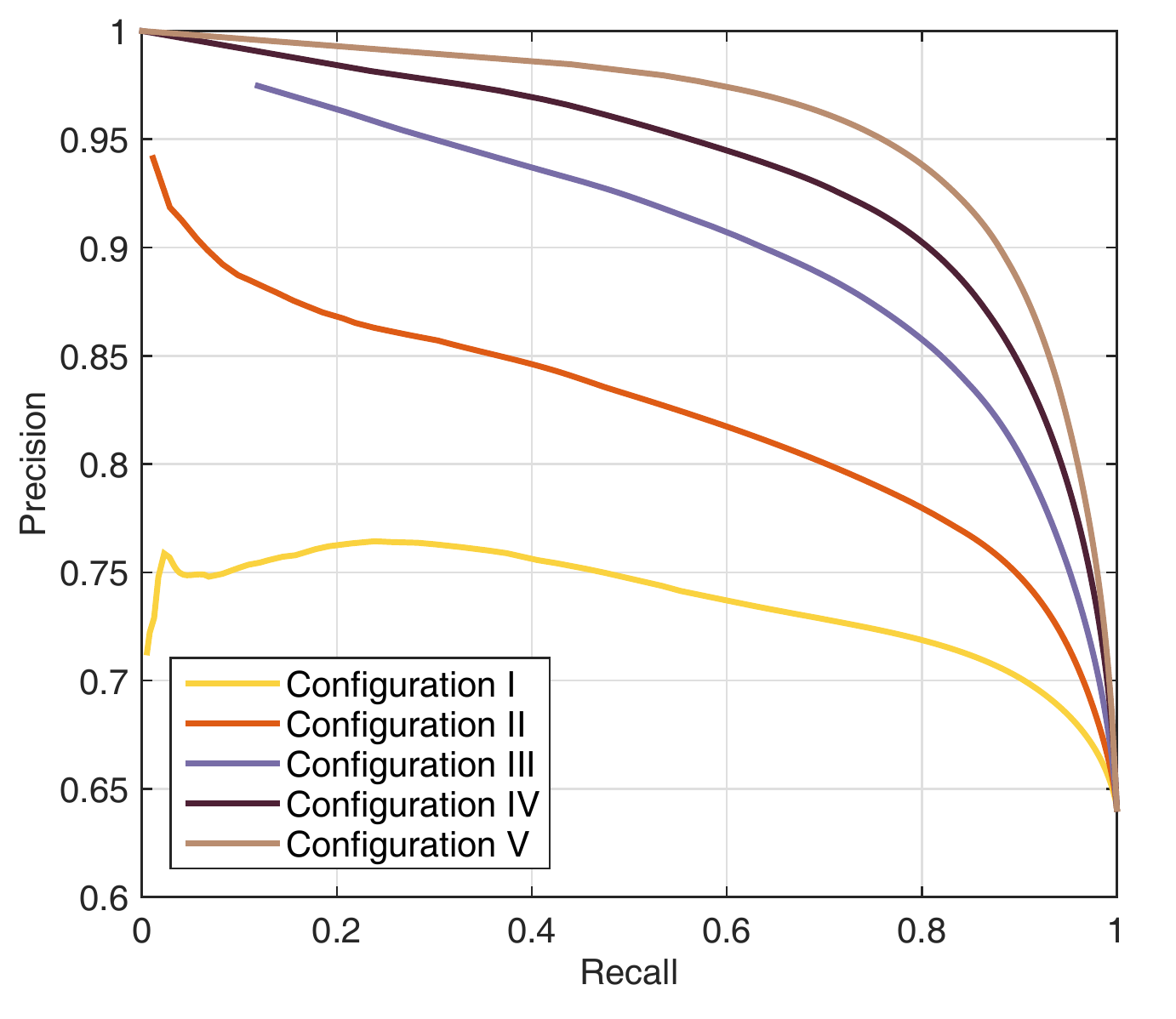}
  \caption{The precision-recall curves of the five configurations trained on $D_o$ and tested
on $D_e$ from the blur detection benchmark~\cite{shi2014discriminative}. Configuration V that favors high-level semantics performs the best at all recall levels compared to the rest shallower configurations. Therefore, it is adopted as the default architecture.}\label{fig:pr_configuration}
\end{figure}

\begin{table}
  \centering
  \caption{Results of the five configurations trained on $D_o$ and tested on $D_e$ from the blur detection benchmark~\cite{shi2014discriminative}}\label{tab:result_configuration}
  \begin{tabular}{l|ccc}
      \toprule
    Algorithm  & ODS & OIS & AP\\
     \hline
     Configuration I& 0.769 & 0.790&0.704 \\
     Configuration II& 0.788 &   0.816&0.772 \\
     Configuration III& 0.815 & 0.850&0.825 \\
     Configuration IV&  0.836 & 0.870&0.855 \\
     \hline
     Configuration V&  {\bf 0.853} & {\bf 0.884}&  {\bf 0.880} \\
     \bottomrule
   \end{tabular}
\end{table}

\begin{figure*}
    \centering
    \captionsetup{justification=centering}
    \subfloat[]{\includegraphics[width=0.14\textwidth]{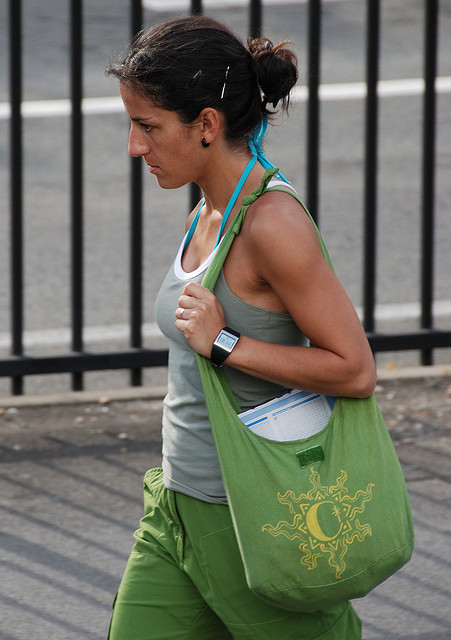}}\hskip.1em
    \subfloat[]{\includegraphics[width=0.14\textwidth]{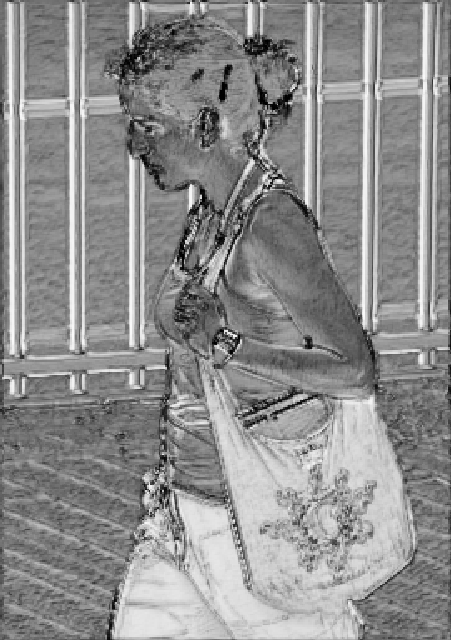}}\hskip.1em
    \subfloat[]{\includegraphics[width=0.14\textwidth]{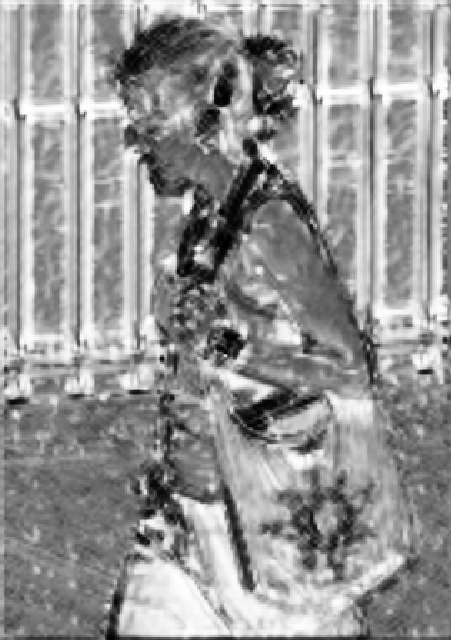}}\hskip.1em
    \subfloat[]{\includegraphics[width=0.14\textwidth]{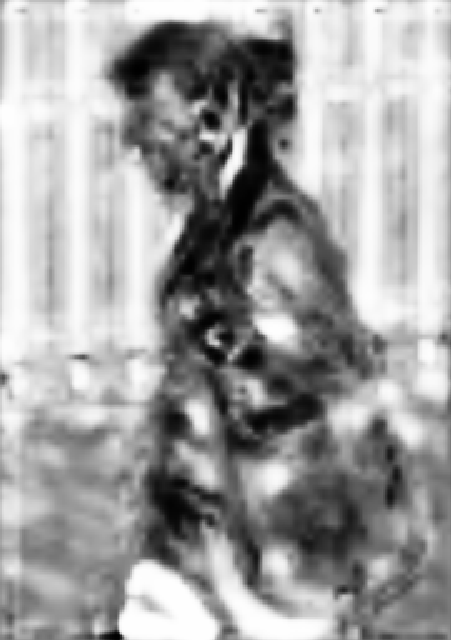}}\hskip.1em
    \subfloat[]{\includegraphics[width=0.14\textwidth]{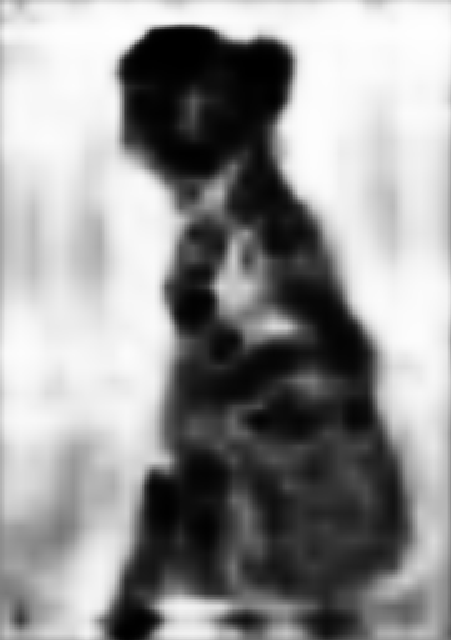}}\hskip.1em
    \subfloat[]{\includegraphics[width=0.14\textwidth]{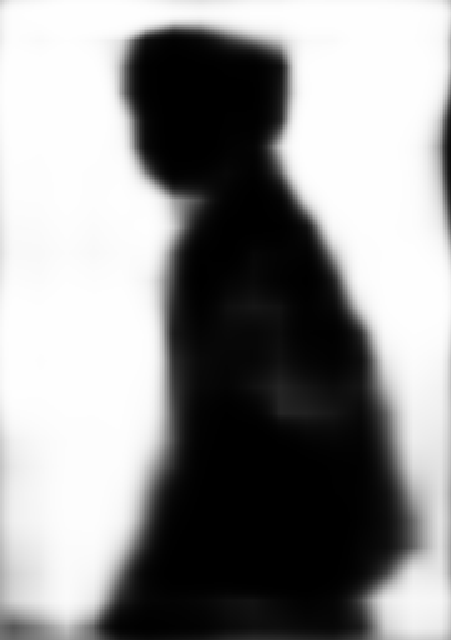}}
    \subfloat[]{\includegraphics[width=0.14\textwidth]{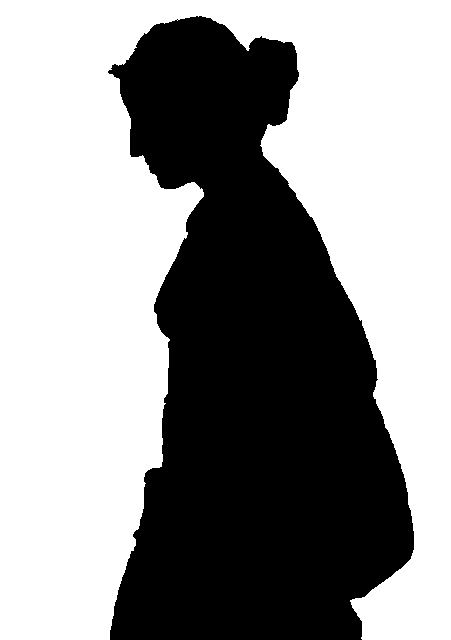}}
    \caption{Comparisons of blur maps generated by the five configurations. (a) Test image from the blur detection benchmark~\cite{shi2014discriminative}. (b) Configuration I. (c) Configuration II. (d) Configuration III. (e) Configuration IV. (f) Configuration V. (g) Ground truth.
 }\label{fig:configMap}
\end{figure*}

\begin{figure}
  \centering
  \includegraphics[width=.5\textwidth]{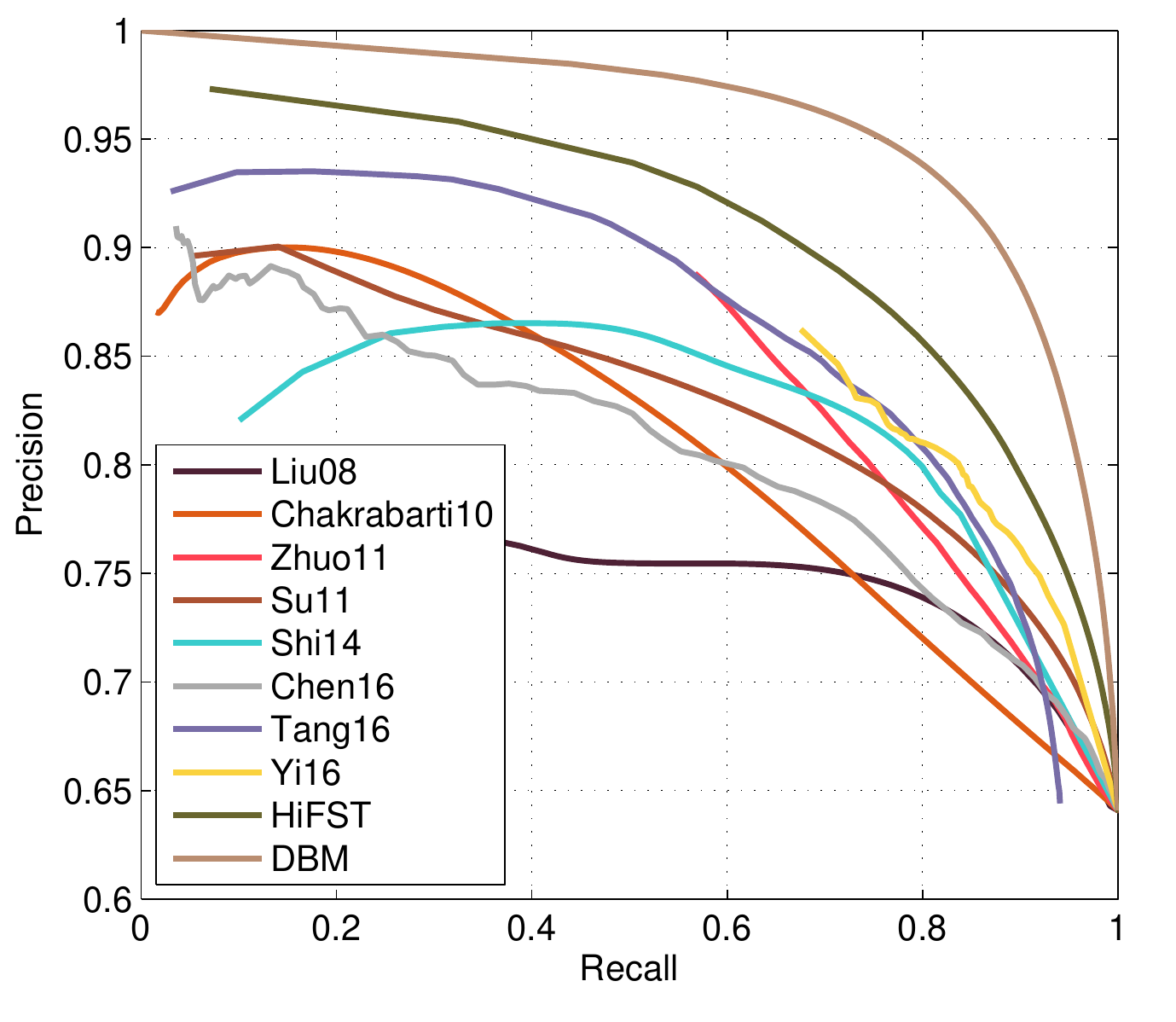}
  \caption{The precision-recall curves trained on $D_o$ and tested on $D_e$. DBM boosts the precisions within the entire recall range, where the improvement can be  as large as $0.2$.}\label{fig:pr_even}
\end{figure}

\subsection{Experimental Results}
\label{subsec:results}

 \subsubsection{Configuration Comparison}\label{subsubsec:config}We first compare the five DBM configurations to investigate the role of spatial information versus semantic information. The quantitative performance is evaluated using the precision-recall curve, which is drawn by concatenating all test images into one vector rather than averaging the curves over all test images. We also summarize the performance using three standard criteria: (1) optimal dataset scale (ODS) F-score obtained by finding an optimal threshold for all images in the dataset; (2) optimal image scale (OIS) F-score by averaging the best F-scores for all images; and (3) average precision (AP) by averaging over all recall levels~\cite{arbelaez2011contour}. We draw the precision-recall curves in Fig.~\ref{fig:pr_configuration}, from which we observe that the performance increases with the depth of configurations at all recall levels. The performance gain is also clear in terms of ODS, OIS, and AP values in Table~\ref{tab:result_configuration}. When the configuration goes deeper, it puts more emphasis on high-level features that contain rich semantic information but does not fully respect the global structure encoded by the spatial information. Therefore, it is clear that semantic information learned by deeper configurations plays a more important role than spatial information, which is not surprising because the blur maps are expected to be relatively uniform, and consist of several clustered and connected regions. This stands out in stark contrast to the edge maps~\cite{xie2015holistically}, where edges scatter across the whole space and spatial information in different scales is essential for accurate edge detection. To take a closer look, we show the blur maps generated by the five architectures in Fig.~\ref{fig:configMap}, from which we can see that shallower configurations can only make use of low-level (gradient-based) features and tend to  mark smooth regions as blurry. Although they have finer spatial information, the generated maps are less relevant to blur perception. By contrast, deeper configurations seem to make decisions based on extracted semantic information and are less affected by the appearances of subjects in the scene. They generate blur maps in closer agreement with the ground truths. In summary, we adopt Configuration V that makes the best use of high-level semantics and delivers the best performance as our default architecture in the rest of the paper.

\subsubsection{Main Results}We next compare DBM with nine existing methods: Liu08~\cite{liu2008image}, Chakrabarti10~\cite{chakrabarti2010analyzing}, Zhuo11~\cite{zhuo2011defocus}, Su11~\cite{su2011blurred}, Shi14~\cite{shi2014discriminative}, Chen16~\cite{chen2016fast}, Tang16~\cite{tang2016spectral}, Yi16~\cite{yi2016lbp}, and HiFST~\cite{alireza2017spatially}. All of them are based on hand-crafted low-level features. The blur maps of each method are either obtained from the original authors or generated by the publicly available implementation with default settings. The precision-recall curves are shown in Fig.~\ref{fig:pr_even}. DBM achieves the highest precisions for all the recall levels, where the improvement can be as large as $0.2$. It is interesting to note that previous methods experience precision drops at low recall levels. This is no surprise because traditional methods tend to give flat regions high blur confidence and misclassify them into blurry regions even with relatively large thresholds. By contrast, DBM automatically learns rich discriminative features, especially high-level semantics, which accurately discriminate flat regions from blurred ones, resulting in nearly perfect precisions at low recall levels. Moreover, DBM exhibits a less steep decline at the middle recall range $[0.2, 0.8]$. This may result from the accurate classification of structures with and without blurring. Table~\ref{tab:result_even} lists the ODS, OIS, and AP results, from which we observe that DBM significantly advances the state-of-the-art by a large margin with an ODS F-score of $0.853$.

\begin{table}
  \centering
  \caption{Results trained on $D_o$ and tested on $D_e$}\label{tab:result_even}
  \begin{tabular}{l|ccc}
      \toprule
    Algorithm  & ODS & OIS & AP\\
     \hline
     Liu08~\cite{liu2008image}& 0.766 & 0.811&0.745 \\
     Chakrabarti10~\cite{chakrabarti2010analyzing}& 0.758 &   0.797&0.757 \\
     Zhuo11~\cite{zhuo2011defocus}& 0.761 & 0.862&0.687 \\
     Su11~\cite{su2011blurred}& 0.782 & 0.822&0.721 \\
     Shi14~\cite{shi2014discriminative}&  0.776 & 0.813&0.843 \\
    Chen16~\cite{chen2016fast}&  0.771 & 0.867&0.823 \\
    Tang16~\cite{tang2016spectral}&   0.765 & 0.864&0.774 \\
     Yi16~\cite{yi2016lbp}&  0.803 & 0.841&0.765 \\
    HiFST~\cite{alireza2017spatially}&  0.813 & 0.851&0.717\\
     \hline
     DBM&  {\bf 0.853} & {\bf 0.884}&  {\bf 0.880} \\
     \bottomrule
   \end{tabular}
\end{table}

\begin{figure*}
    \centering
    \captionsetup{justification=centering}

    \subfloat{\includegraphics[width=0.19\textwidth]{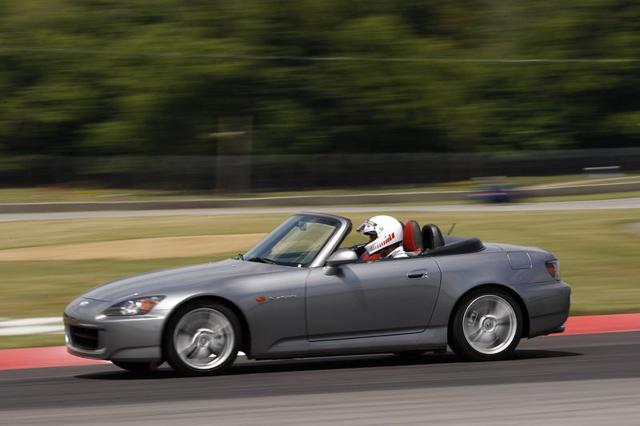}}\hskip.2em
    \subfloat{\includegraphics[width=0.19\textwidth]{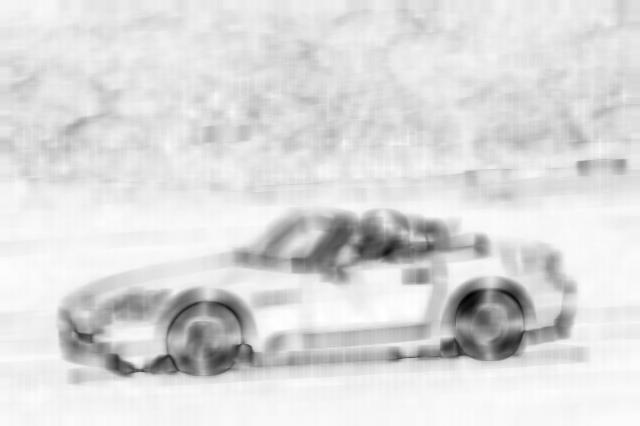}}\hskip.2em
    \subfloat{\includegraphics[width=0.19\textwidth]{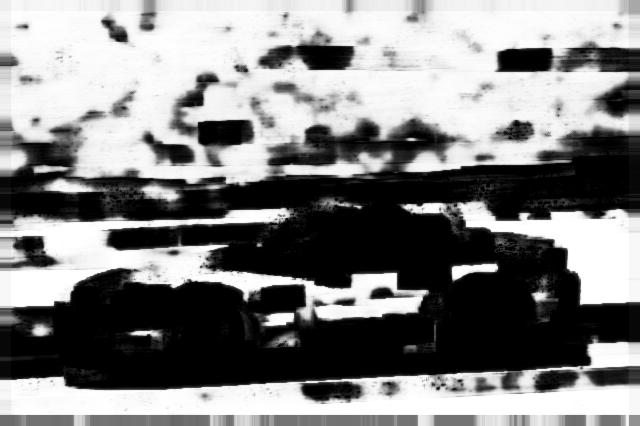}}\hskip.2em
    \subfloat{\includegraphics[width=0.19\textwidth]{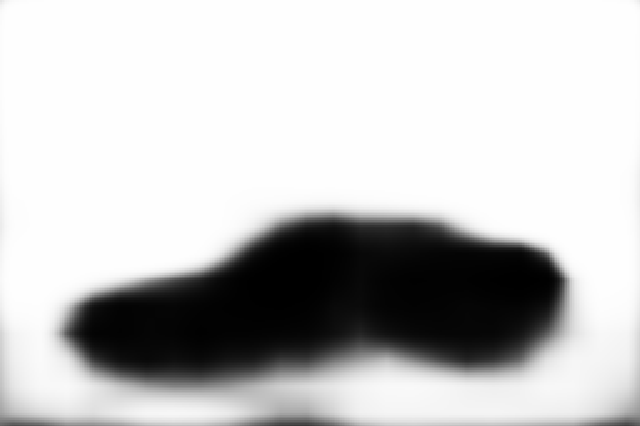}}\hskip.2em
    \subfloat{\includegraphics[width=0.19\textwidth]{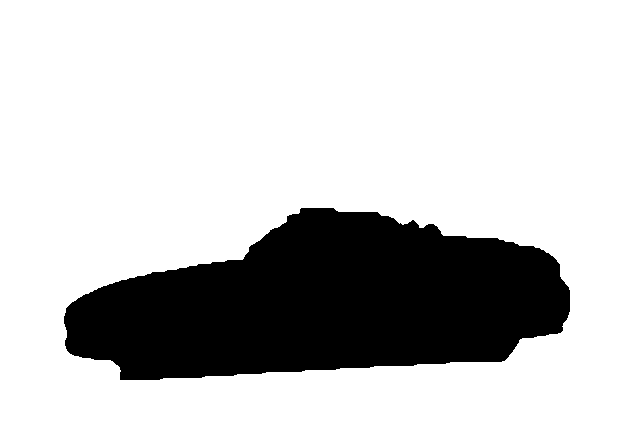}}
    \vspace{-6pt}
    \subfloat{\includegraphics[width=0.19\textwidth]{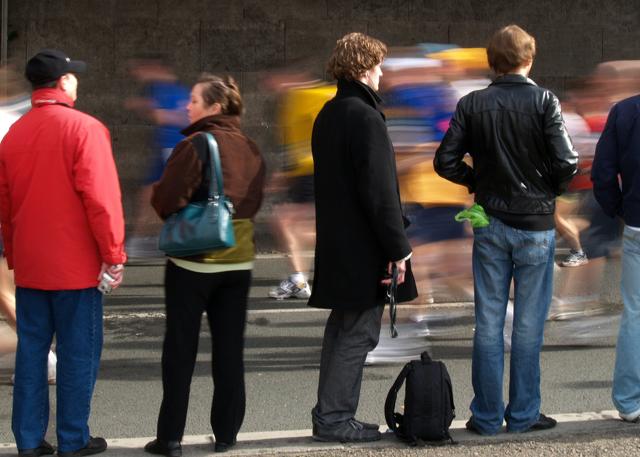}}\hskip.2em
    \subfloat{\includegraphics[width=0.19\textwidth]{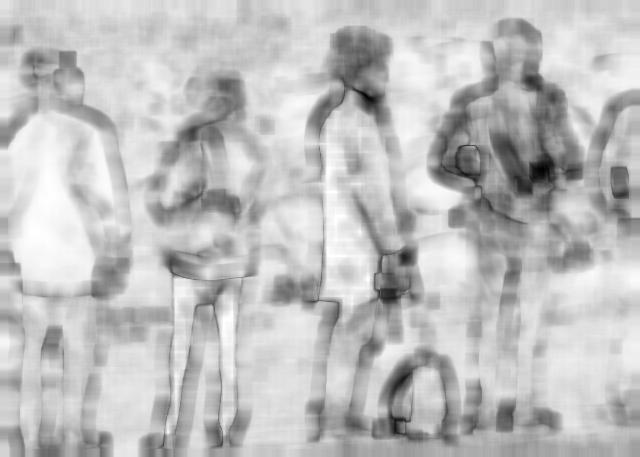}}\hskip.2em
    \subfloat{\includegraphics[width=0.19\textwidth]{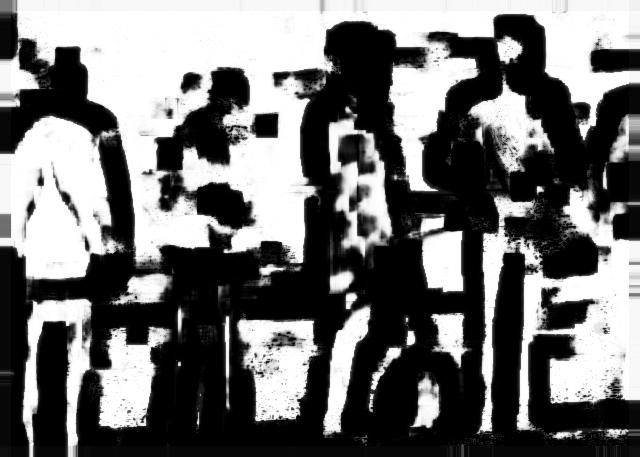}}\hskip.2em
    \subfloat{\includegraphics[width=0.19\textwidth]{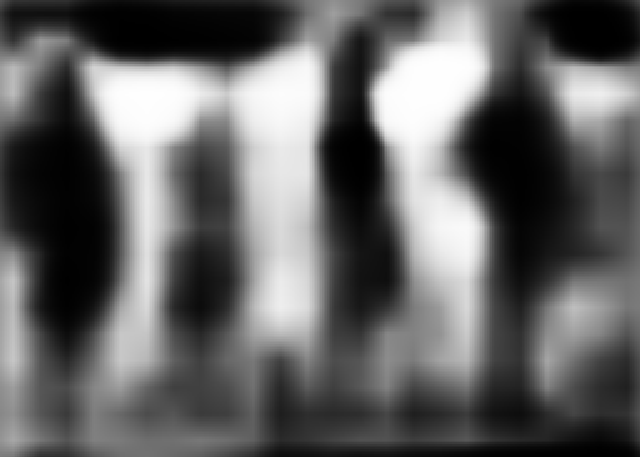}}\hskip.2em
    \subfloat{\includegraphics[width=0.19\textwidth]{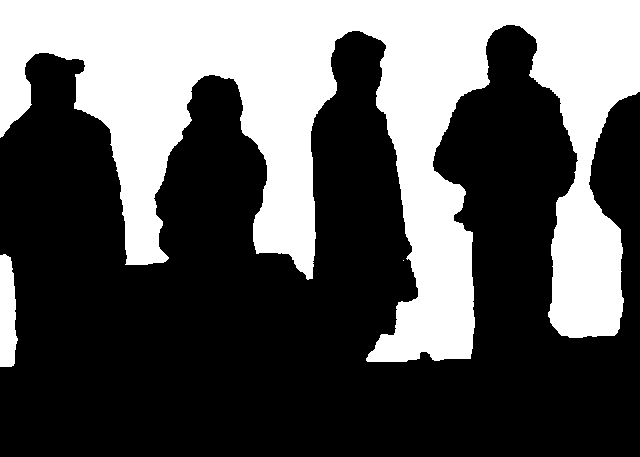}}
    % \vspace{-6pt}
    % \subfloat{\includegraphics[width=0.19\textwidth]{figs/motion0088}}\hskip.2em
    % \subfloat{\includegraphics[width=0.19\textwidth]{figs/motion0088_su}}\hskip.2em
    % \subfloat{\includegraphics[width=0.19\textwidth]{figs/motion0088_shi}}\hskip.2em
    % \subfloat{\includegraphics[width=0.19\textwidth]{figs/motion0088_us}}\hskip.2em
    % \subfloat{\includegraphics[width=0.19\textwidth]{figs/motion0088_gt}}
    \vspace{-6pt}
    \subfloat{\includegraphics[width=0.19\textwidth]{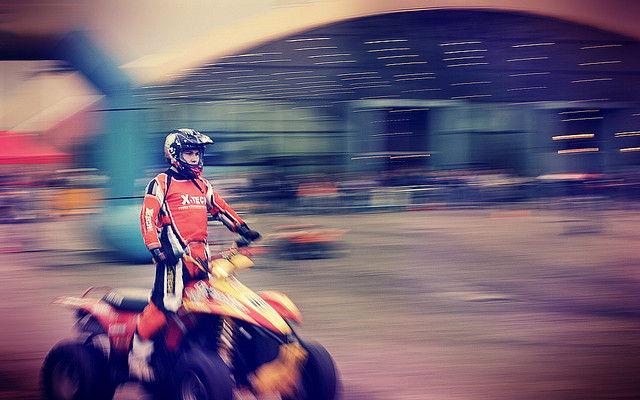}}\hskip.2em
    \subfloat{\includegraphics[width=0.19\textwidth]{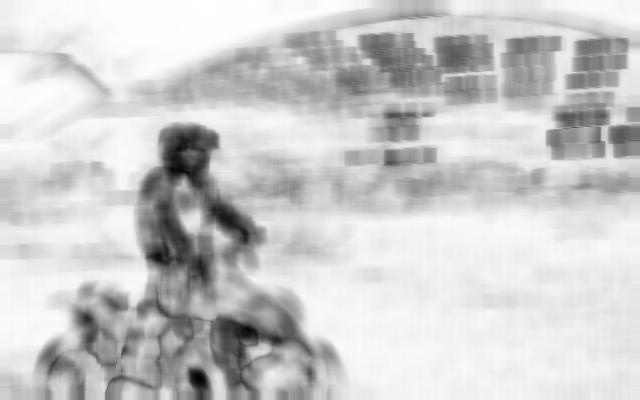}}\hskip.2em
    \subfloat{\includegraphics[width=0.19\textwidth]{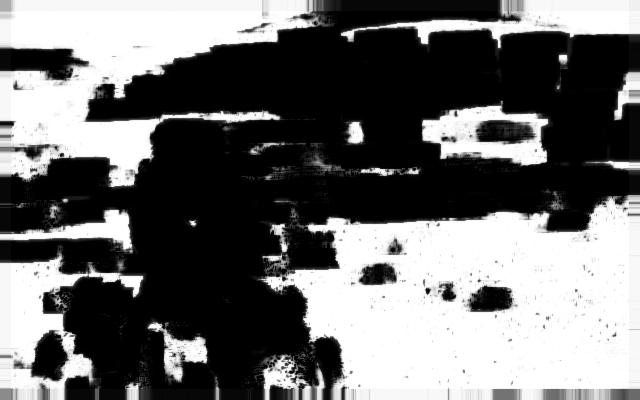}}\hskip.2em
    \subfloat{\includegraphics[width=0.19\textwidth]{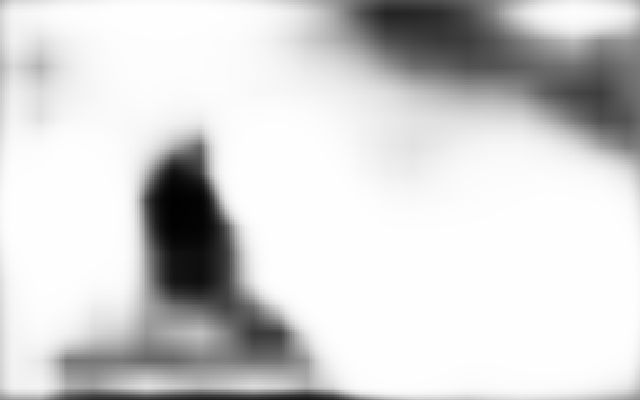}}\hskip.2em
    \subfloat{\includegraphics[width=0.19\textwidth]{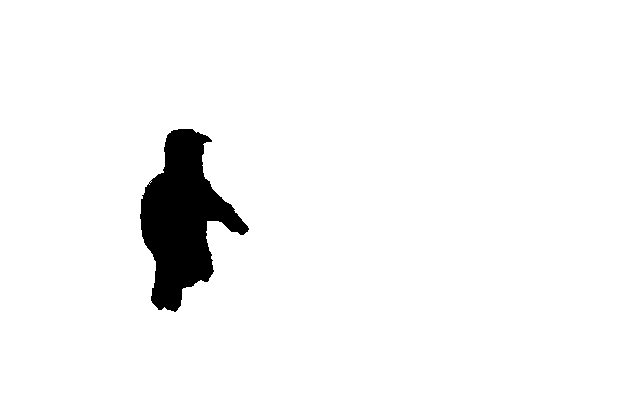}}
    % \vspace{-6pt}
    % \subfloat{\includegraphics[width=0.19\textwidth]{figs/motion0018}}\hskip.2em
    % \subfloat{\includegraphics[width=0.19\textwidth]{figs/motion0018_su}}\hskip.2em
    % \subfloat{\includegraphics[width=0.19\textwidth]{figs/motion0018_shi}}\hskip.2em
    % \subfloat{\includegraphics[width=0.19\textwidth]{figs/motion0018_us}}\hskip.2em
    % \subfloat{\includegraphics[width=0.19\textwidth]{figs/motion0018_gt}}
    \vspace{-6pt}
    \subfloat{\includegraphics[width=0.19\textwidth]{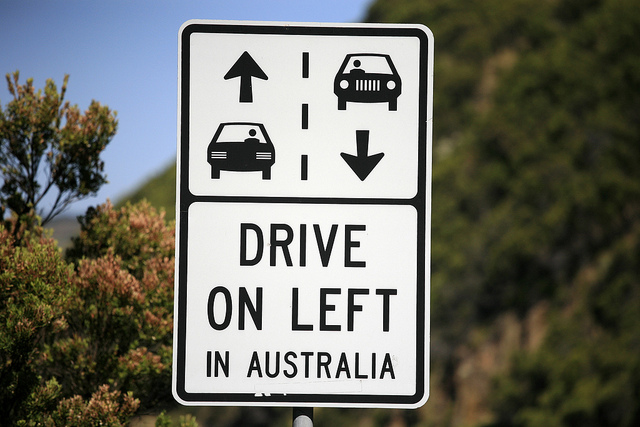}}\hskip.2em
    \subfloat{\includegraphics[width=0.19\textwidth]{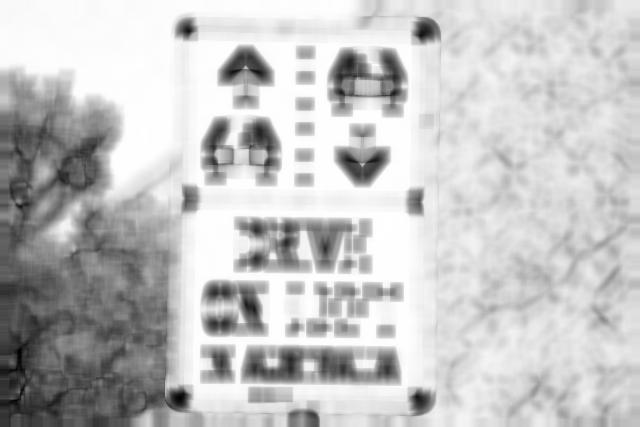}}\hskip.2em
    \subfloat{\includegraphics[width=0.19\textwidth]{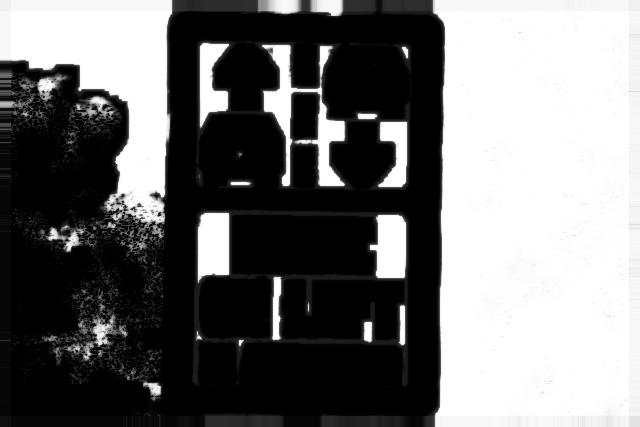}}\hskip.2em
    \subfloat{\includegraphics[width=0.19\textwidth]{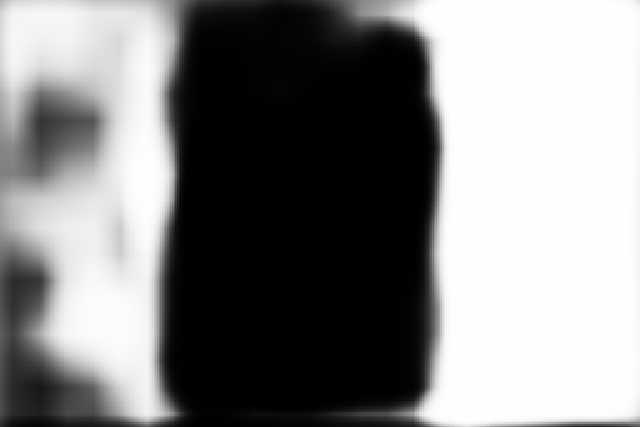}}\hskip.2em
    \subfloat{\includegraphics[width=0.19\textwidth]{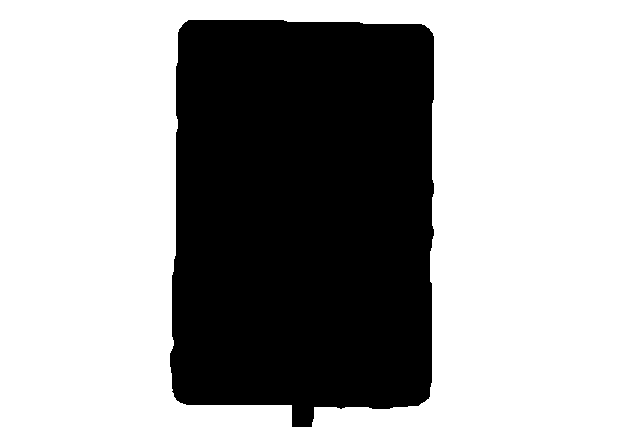}}
    \vspace{-6pt}
%    \subfloat{\includegraphics[width=0.19\textwidth]{figs/out_of_focus0568}}\hskip.2em
%    \subfloat{\includegraphics[width=0.19\textwidth]{figs/out_of_focus0568_su}}\hskip.2em
%    \subfloat{\includegraphics[width=0.19\textwidth]{figs/out_of_focus0568_shi}}\hskip.2em
%    \subfloat{\includegraphics[width=0.19\textwidth]{figs/out_of_focus0568_us}}\hskip.2em
%    \subfloat{\includegraphics[width=0.19\textwidth]{figs/out_of_focus0568_gt}}
%    \vspace{-9pt}
    \subfloat{\includegraphics[width=0.19\textwidth]{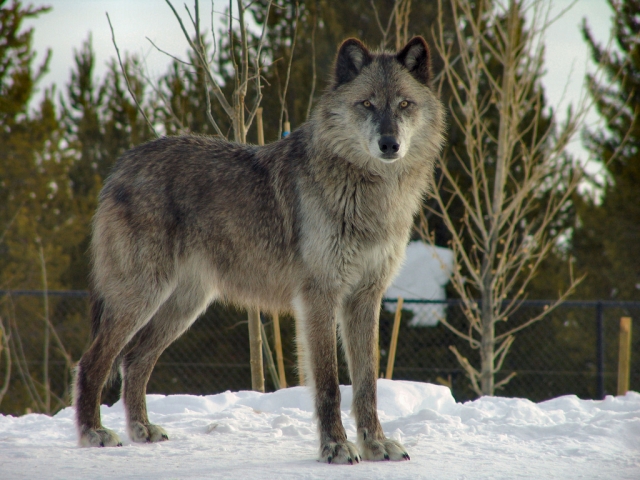}}\hskip.2em
    \subfloat{\includegraphics[width=0.19\textwidth]{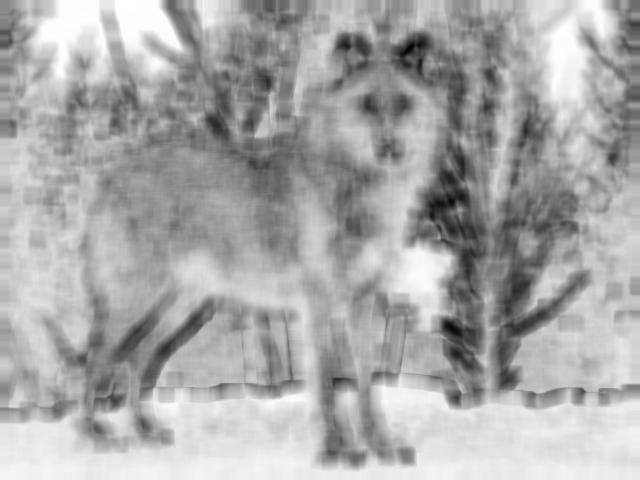}}\hskip.2em
    \subfloat{\includegraphics[width=0.19\textwidth]{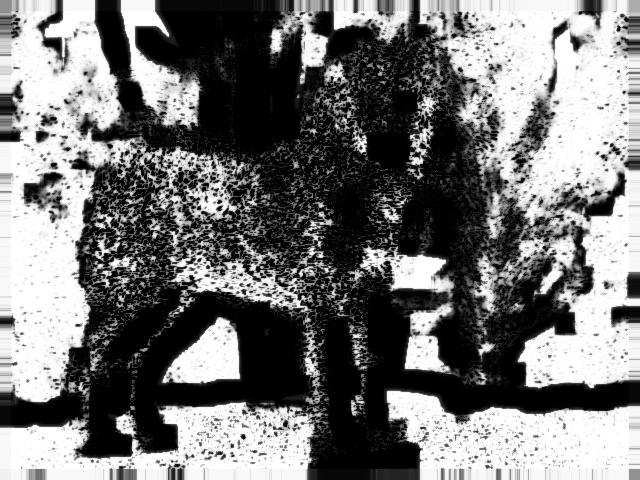}}\hskip.2em
    \subfloat{\includegraphics[width=0.19\textwidth]{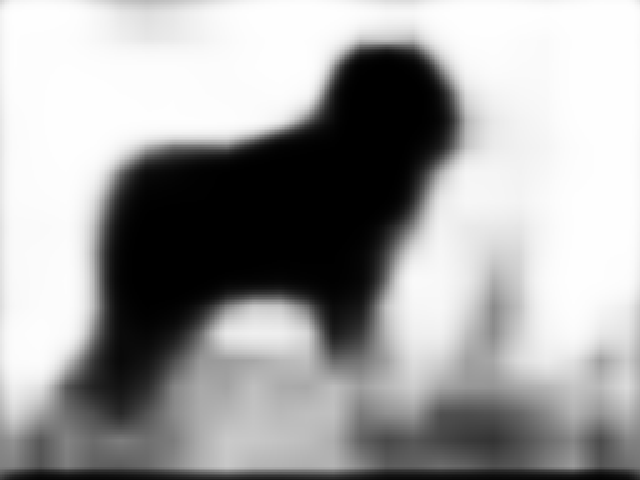}}\hskip.2em
    \subfloat{\includegraphics[width=0.19\textwidth]{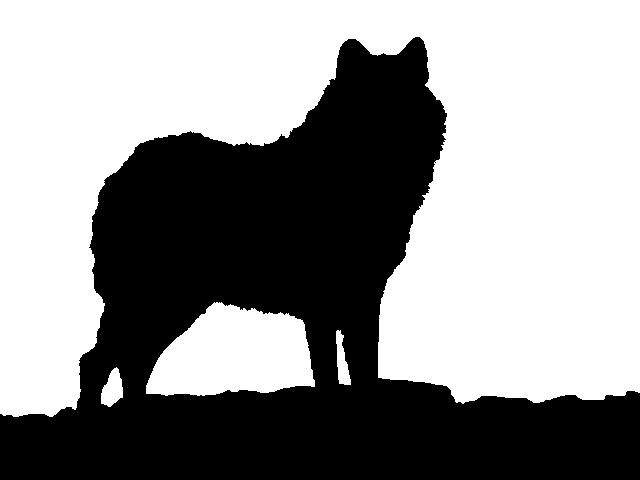}}
    \vspace{-6pt}
    \addtocounter{subfigure}{-25}
    \subfloat[]
    {\includegraphics[width=0.19\textwidth]{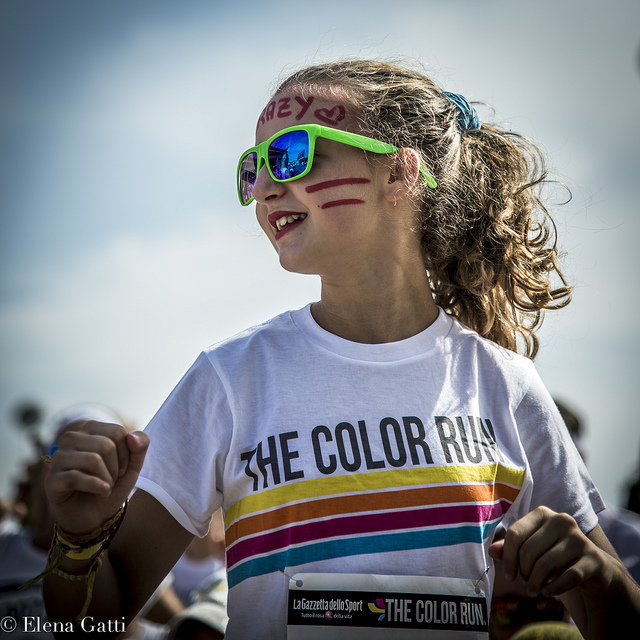}}\hskip.2em
    \subfloat[]{\includegraphics[width=0.19\textwidth]{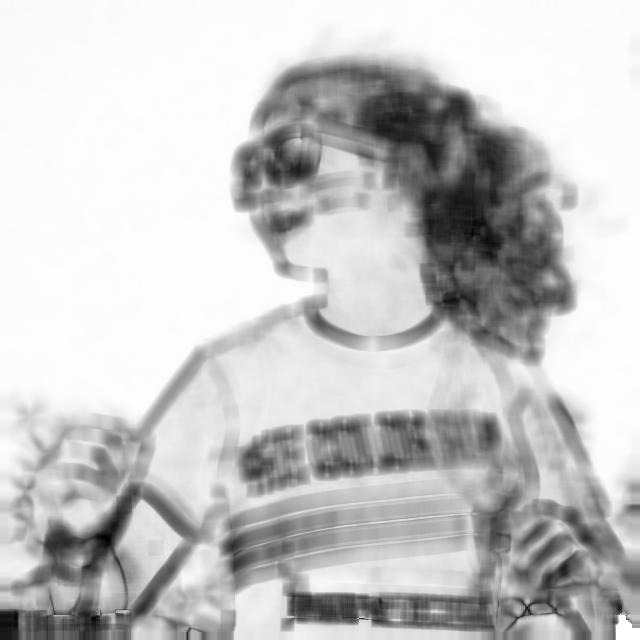}}\hskip.2em
    \subfloat[]{\includegraphics[width=0.19\textwidth]{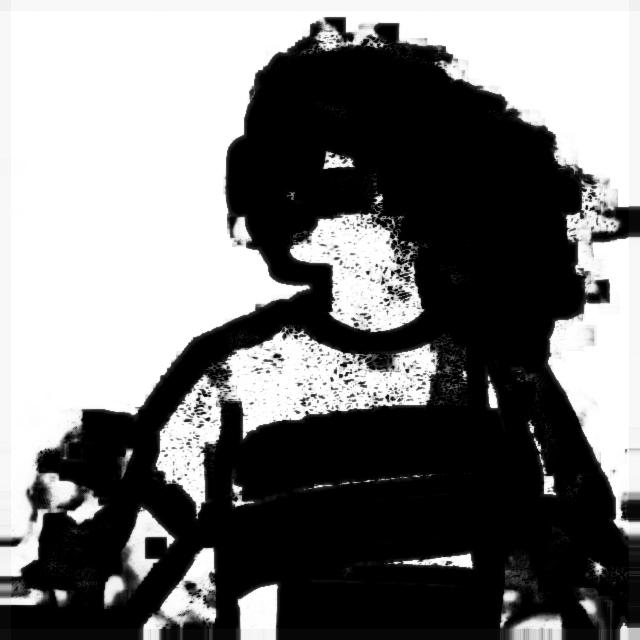}}\hskip.2em
    \subfloat[]{\includegraphics[width=0.19\textwidth]{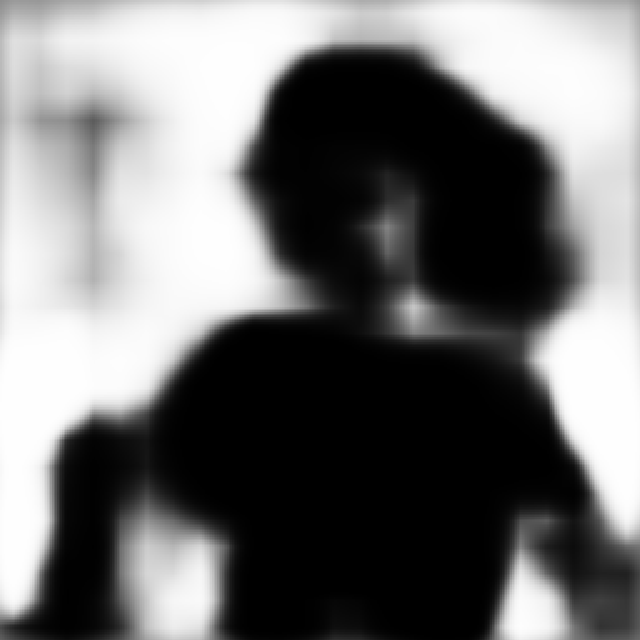}}\hskip.2em
    \subfloat[]{\includegraphics[width=0.19\textwidth]{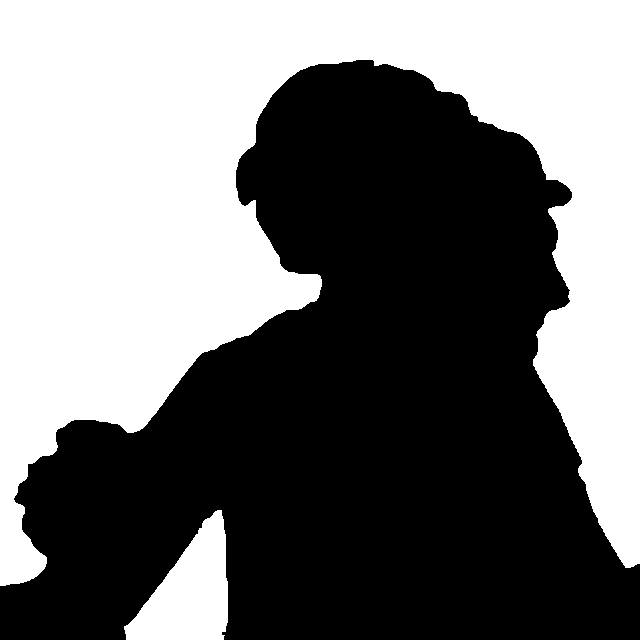}}

    \caption{Representative blur mapping results on the blur detection benchmark~\cite{shi2014discriminative}. (a) Test images. (b) Su11~\cite{su2011blurred}. (c)  Shi14~\cite{shi2014discriminative}. (d) DBM. (e) Ground truths. The proposed DBM shows a clear advantage in terms of accuracy over Su11~\cite{su2011blurred} and Shi14~\cite{shi2014discriminative}, and is more consistent with the ground truths. }\label{fig:vd}
\end{figure*}

To better investigate the effectiveness of DBM at detecting local blur, we show some blur maps generated by DBM and compare them with those by two representative methods Su11~\cite{su2011blurred} and Shi14~\cite{shi2014discriminative} in Fig.~\ref{fig:vd}. DBM is able to robustly detect local blur from  complex foreground and background. First, it well handles  blur regions across different scales from the small motorcycle man (in the $3$-th row) to the big girl (in the $6$-th row). Second, it is capable of identifying  flat regions such as the car body (in the first row), clothes (in the $2$-nd and $6$-th rows), and the road sign (in the $4$-th row) as non-blurry. Third, it is barely affected by strong structures after blurring and labels those regions correctly. All of these stand in stark contrast with previous methods, which mix flat and blurry regions with high probability, and are severely biased by strong structures after blurring. Moreover, DBM labels images with high confidence. Nearly $40\%$ pixels in the test images have predicted values either larger than $0.9$ (blurry) or smaller than $0.1$ (non-blurry).

\begin{figure*}[t]
    \centering
    \captionsetup{justification=centering}
    \subfloat[]{\includegraphics[width=0.14\textwidth]{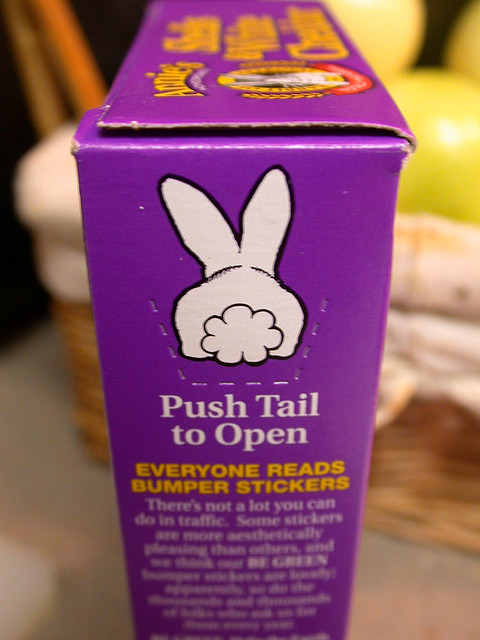}}\hskip.1em
    \subfloat[]{\includegraphics[width=0.14\textwidth]{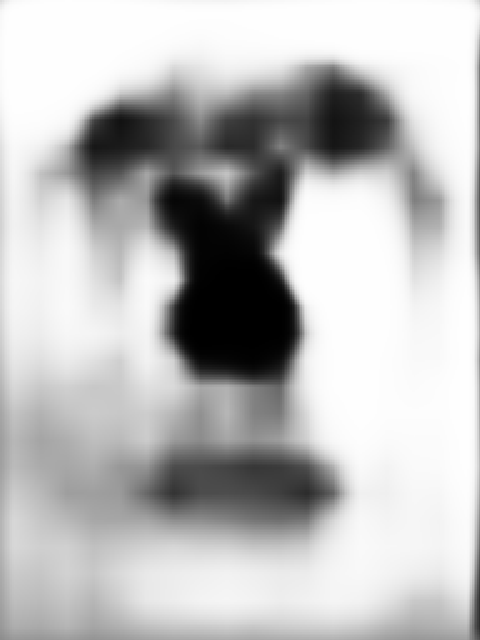}}\hskip.1em
    \subfloat[]{\includegraphics[width=0.14\textwidth]{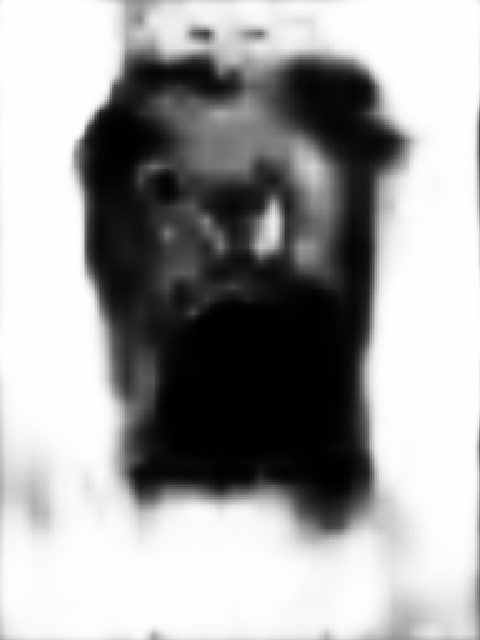}}\hskip.1em
    \subfloat[]{\includegraphics[width=0.14\textwidth]{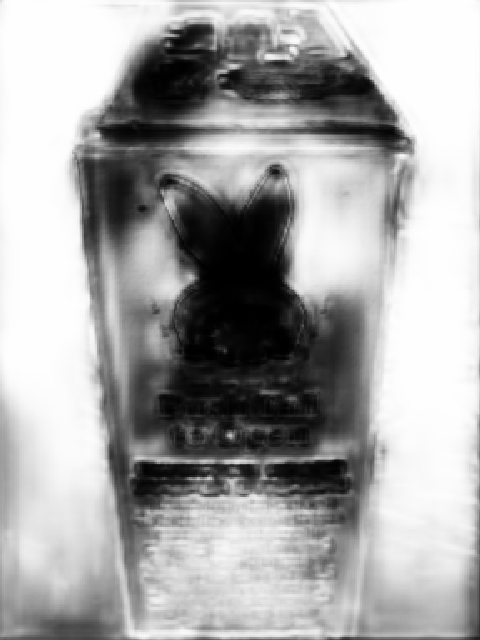}}\hskip.1em
    \subfloat[]{\includegraphics[width=0.14\textwidth]{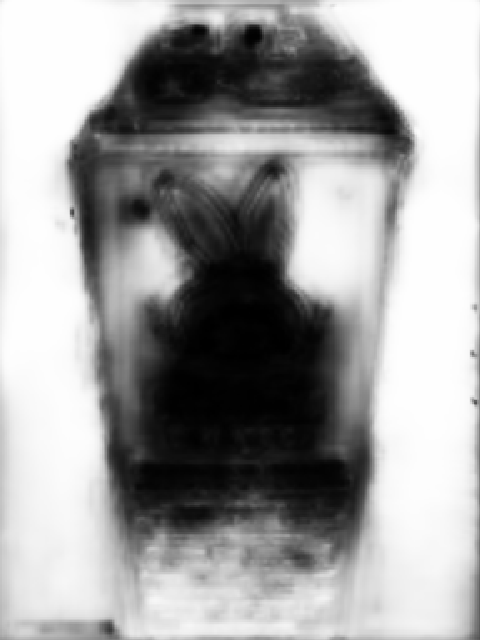}}\hskip.1em
    \subfloat[]{\includegraphics[width=0.14\textwidth]{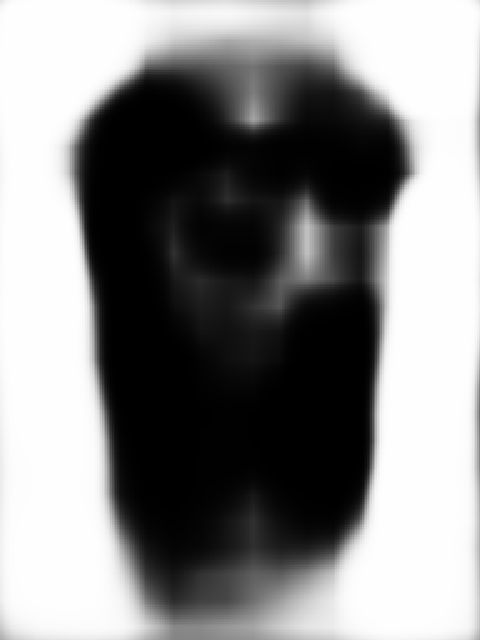}}
    \subfloat[]{\includegraphics[width=0.14\textwidth]{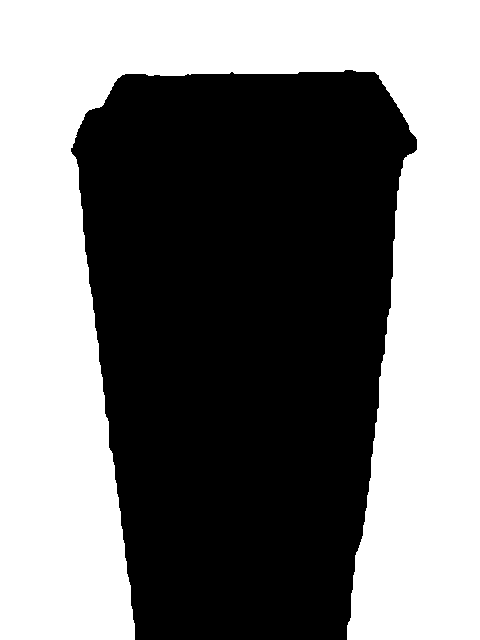}}
    \caption{Comparisons of blur maps generated by the variants of DBM. (a) Test image from the blur detection benchmark~\cite{shi2014discriminative}. (b) Training from scratch. (c) Training with FCNs and skip layers~\cite{long2015fully} (FCN-8s). (d) Training with weighted fusion only~\cite{xie2015holistically}. (e) Training with weighted fusion and deep supervision~\cite{xie2015holistically} (DSN). (f) DBM. (g) Ground truth.
 }\label{fig:hl}
\end{figure*}

\subsubsection{Importance of High-Level Semantics}\label{subsubsec:high}Besides analyzing the importance of high-level semantics via architectures with different depths, we conduct another series of experiments to show that the learned high-level features indeed play a crucial role in our blur mapper. We first train DBM from scratch without using semantically meaningful initializations. The hyper-parameters are manually optimized to guarantee the best performance. The results shown in the first row of Table~\ref{tab:analysis} are unsatisfactory, which is expected since bad initializations can stall learning due to the instability of gradients in deep nets. By contrast, a more informative initialization with respect to the blur mapping task (in this case from semantic segmentation) is likely to guide SGD to find better local minima and results in a more meaningful blur map (Fig.~\ref{fig:hl} (f)).

We then investigate more advanced network architectures that make better use of low-level features at shallower layers, including FCNs with skip layers (FCN-8s)~\cite{long2015fully}, weighted fusion of side outputs~\cite{xie2015holistically}, and weighted fusion of side outputs and deep supervision~\cite{xie2015holistically} (DSN). The results are shown in Table~\ref{tab:analysis} and Fig.~\ref{fig:hl}. We observe that although incorporating low-level features produces blur maps with somewhat finer spatial information (similar to what we have observed in Fig.~\ref{fig:configMap}), it voids the benefits of high-level features and results in erroneous and non-uniform blur assignments. This is expected because low-level features mainly contain edge information of an input image and do not help blur detection much. FCN-8s~\cite{long2015fully} that treats low-level and high-level features with equal importance impairs  performance the most. The weighted fusion scheme without deep supervision learns to assign importance weights to the side outputs. It turns out that the side outputs generated by deeper convolutional layers are weighted heavier than those by shallower layers. Specifically, the learned fusion weights of the five side outputs from shallow to deep layers are $[0.052, 0.149, 0.160, 0.273, 0.384]$, respectively. We observe slight performance improvement over FCN-8s. The weighted fusion scheme with deep supervision directly regularizes low-level features using the ground truth and delivers slightly better performance in terms of ODS and OIS than DBM. In summary, DBM that solely interpolates from high-level feature maps achieves comparable performance to its most sophisticated variant DSN and ranks the best in terms of AP. The blur maps by DBM are more reasonable and closer to the ground truths perceptually. These manifest the central role of high-level semantics in local blur mapping.
\begin{table}
  \centering
  \caption{Comparing DBM with its variants to identify the role of high-level semantics}\label{tab:analysis}
  \begin{tabular}{l|ccc}
      \toprule
      & ODS & OIS & AP\\
     \hline
     Training from scratch& 0.833 & 0.876&0.856 \\
     FCN-8s& 0.840 & 0.874&0.847 \\
     Fusion (w/o deep supervision)&0.844 &  0.877 & 0.865 \\

     Fusion (with deep supervision)& {\bf 0.854} & {\bf 0.889} &0.876 \\
     \hline
    DBM&  0.853 & 0.884 &  {\bf 0.880} \\
     \bottomrule
   \end{tabular}
\end{table}
\begin{table}
  \centering
  \caption{Results trained on $D_e$ and tested on $D_o$}\label{tab:result_odd}
  \begin{tabular}{l|ccc}
      \toprule
    Algorithm & ODS & OIS & AP\\
     \hline
     Liu08~\cite{liu2008image}& 0.753 & 0.803&0.749 \\
     Chakrabarti10~\cite{chakrabarti2010analyzing}& 0.741 &     0.788&0.741 \\
     Zhuo11~\cite{zhuo2011defocus}& 0.746 & 0.853&0.676 \\
     Su11~\cite{su2011blurred}& 0.775 & 0.814&0.712 \\
     Shi14~\cite{shi2014discriminative}&  0.765 & 0.804&0.831 \\
     Chen16~\cite{chen2016fast}&  0.752& 0.859&0.823 \\
     Tang16~\cite{tang2016spectral}&  0.746 & 0.846& 0.755 \\
     Yi16~\cite{yi2016lbp}&  0.786 & 0.829&0.741 \\
     HiFST~\cite{alireza2017spatially}&   0.804 & 0.841&0.706\\
     
     \hline
     DBM&  {\bf 0.852} & {\bf 0.885}& {\bf 0.876} \\
     \bottomrule
   \end{tabular}
\end{table}

\begin{table*}[t]
  \centering
  \caption{Average execution time in seconds on $10$ images of size $384\times 384\times3$ from the blur detection benchmark~\cite{shi2014discriminative}}\label{tab:time}
  \begin{tabular}{l|ccccccccc}
      \toprule
     \multirow{2}{*}{Algorithm}  &Chakrabarti10&Zhuo11&Su11 &Chen16&Tang16& Yi16&HiFST & DBM  & DBM \\
     
     &\cite{chakrabarti2010analyzing}& \cite{zhuo2011defocus}& \cite{su2011blurred}&\cite{chen2016fast}&\cite{tang2016spectral}&\cite{yi2016lbp}&\cite{alireza2017spatially}&(CPU)&(GPU)\\
     \hline
     Time (s)&0.7$\,\pm\,$0.3& 11.0$\,\pm\,$0.2& 5.1$\,\pm\,$0.1& 1.0$\,\pm\,$0.1&1.4$\,\pm\,$0.2&20.1$\,\pm\,$2.8& 99.7$\,\pm\,$1.4&1.8$\,\pm\,$0.2&0.027$\,\pm\,$0.004\\
 
     \bottomrule
   \end{tabular}
\end{table*}

\begin{figure}
  \centering
  \includegraphics[width=.5\textwidth]{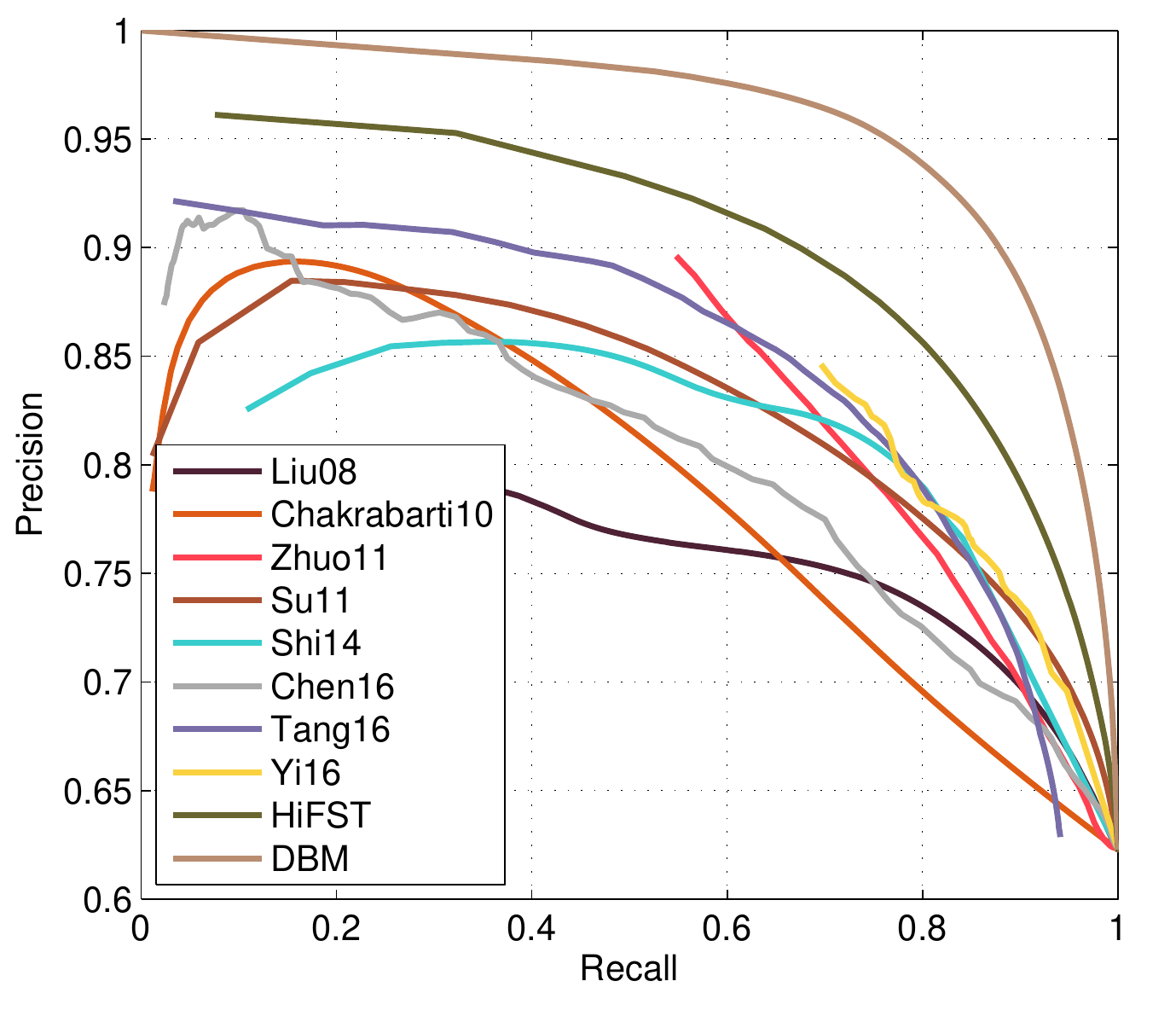}
  \caption{The precision-recall curves trained on $D_e$ and tested on $D_o$. DBM achieves similar superior performance when using $D_o$ for training, indicating its independence of specific training sets.}\label{fig:pr_odd}
\end{figure}

\begin{figure}
    \centering
    \captionsetup{justification=centering}

    \subfloat{\includegraphics[width=0.23\textwidth]{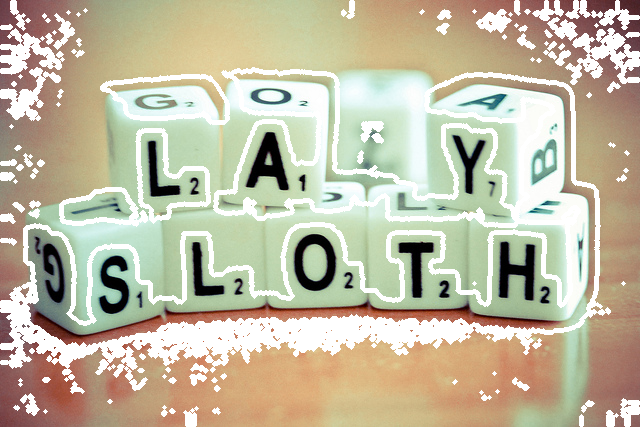}}\hskip.2em
    \subfloat{\includegraphics[width=0.23\textwidth]{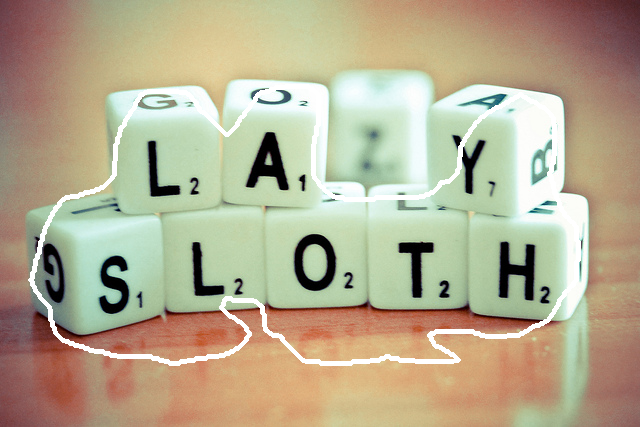}}

    \vspace{-.9pt}
    \addtocounter{subfigure}{-2}
    \subfloat[]
    {\includegraphics[width=0.23\textwidth]{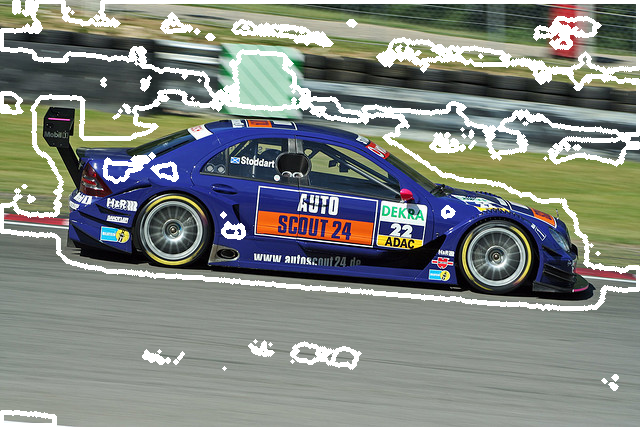}}\hskip.2em
    \subfloat[]{\includegraphics[width=0.23\textwidth]{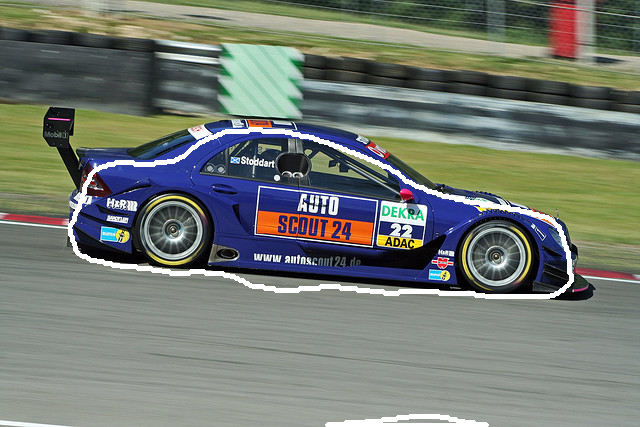}}

    \caption{The blur region segmentation results. (a) Shi14~\cite{shi2014discriminative}. (b) DBM.}\label{fig:blurSeg}
\end{figure}

\subsubsection{Independence of Training Data}It is important to verify the generalizability of DBM by showing that it is  independent of specific training sets. We therefore switch the training and test sets in our setting. In other words, we train DBM on $D_e$ and test on $D_o$. We observe in the Fig.~\ref{fig:pr_odd} and Table~\ref{tab:result_odd} that similar superior performance has been achieved in terms of the precision-recall curve, ODS, OIS, and AP. This verifies that DBM does not rely on any specific training set as long as the set is diverse to cover various natural scenes and  causes of blur.

\subsubsection{More Training Data}Deep learning algorithms have dominated many computer vision tasks, at least in part due to the availability of large amounts of labeled data for training. However, in local blur mapping, we are limited by the number of training images available in the existing benchmark. Here we want to explore whether more training data with novel content further benefit DBM. To do this, we randomly sample $400$ images from $D_e$, incorporate them into $D_o$, and test DBM on the remaining $100$ images. The result averaged over $5$ such trials is reported. We observe that by adding more training images, performance improves from $\text{ODS} = 0.862$ to $\text{ODS} = 0.869$. This indicates that we may further boost the performance and enhance the robustness of DBM
by training it with a larger dataset.

\subsubsection{Running Time}We compare the execution time of DBM with existing methods using $10$ images of size $384\times384\times3$ on a computer with $3.4$GHz CPU and $16$G RAM. From Table~\ref{tab:time}, we see that DBM keeps a good balance between
prediction performance and computational complexity. When the GPU mode is activated (we adopt an NVIDIA
GTX Titan X GPU), DBM runs significantly faster than existing methods, enabling real-time applications.

In summary, we have empirically shown that a linear cascaded FCN-based DBM that exploits high-level semantics delivers superior performance in local blur mapping. The low-level features that better encode gradient and spatial information are less relevant to this task. 
% This verifies our new perspective on blur perception that high-level semantics matter.

%-------------------------------------------------------------------------

\subsection{Applications}
In this subsection, we explore three potential applications that benefit from the blur maps generated by DBM: 1) blur region segmentation, 2) blur degree estimation, and 3) blur magnification.

\subsubsection{Blur Region Segmentation} The goal of image segmentation is to partition an image into multiple regions, which is perceptually more meaningful and easier to analyze~\cite{shi2000normalized}. It is difficult for automatic segmentation algorithms to work well on all images.  Therefore, many interactive image segmentation tools are proposed, which require users to manually create a mask to roughly indicate what parts belong to foreground and background. The blur map produced by DBM provides a useful mask to initialize segmentation without human intervention. Here we adopt GrabCut~\cite{rother2004grabcut}, a popular interactive image segmentation method based on graph cuts, and set pixels with blur confidence $[0,0.1)$, $[0.1, 0.5)$, $[0.5,0.9)$, and $[0.9,1]$ as foreground, probable foreground, probable background, and background, respectively. The implementation we use is based on OpenCV version 3.2 with default settings. We compare our results with Shi14~\cite{shi2014discriminative} in Fig.~\ref{fig:blurSeg} and observe that DBM does a better job in segmenting images into blur and clear regions. By contrast, Shi14~\cite{shi2014discriminative} mislabels flat regions in the foreground and structures with blurring in the background, and segments images into non-connected parts.

\begin{figure*}
    \centering
    \captionsetup{justification=centering}

    \subfloat
    {\includegraphics[width=0.16\textwidth]{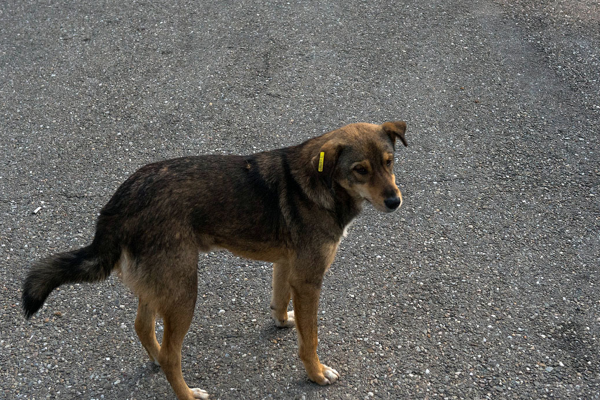}}\hskip.2em
    \subfloat{\includegraphics[width=0.16\textwidth]{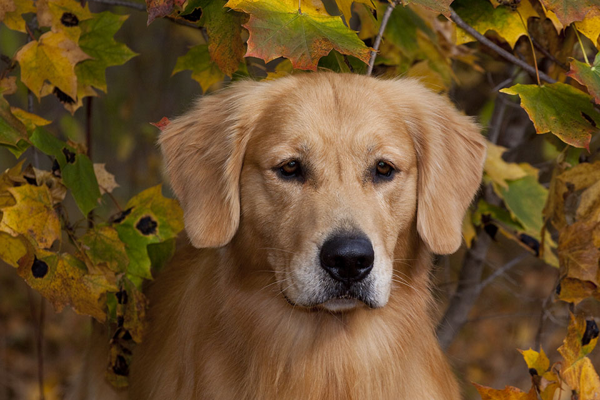}}\hskip.2em
    \subfloat{\includegraphics[width=0.16\textwidth]{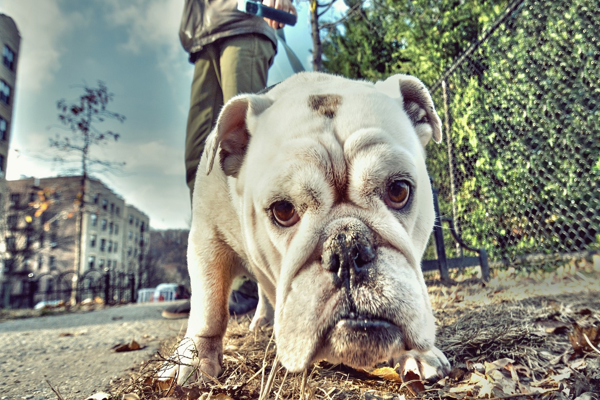}}\hskip.2em
    \subfloat{\includegraphics[width=0.16\textwidth]{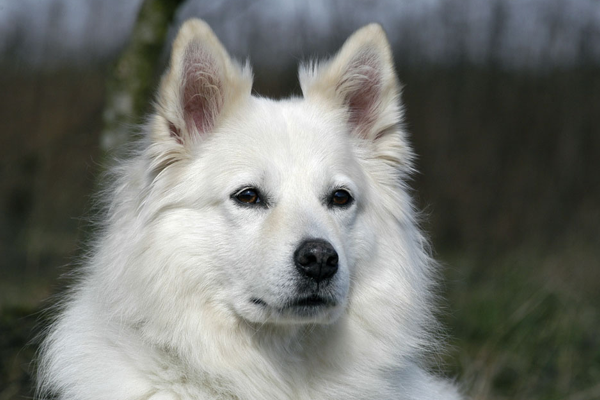}}\hskip.2em
    \subfloat{\includegraphics[width=0.16\textwidth]{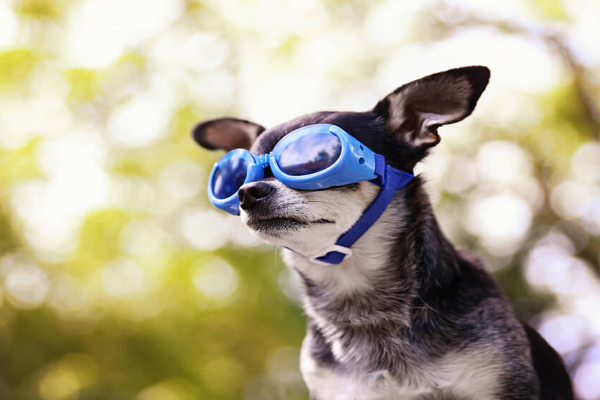}}\hskip.2em
    \subfloat{\includegraphics[width=0.16\textwidth]{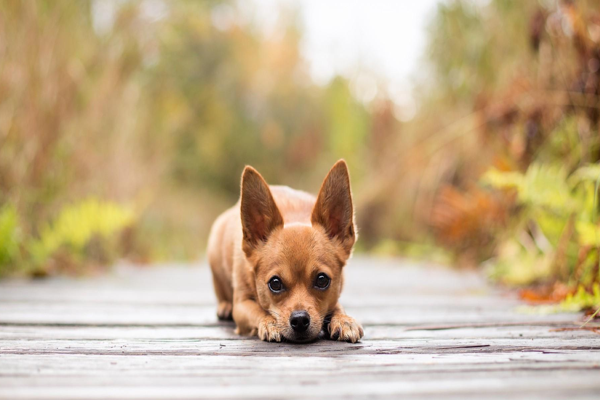}}
    \vspace{2pt}
    \addtocounter{subfigure}{-6}
    \subfloat[]
    {\includegraphics[width=0.16\textwidth]{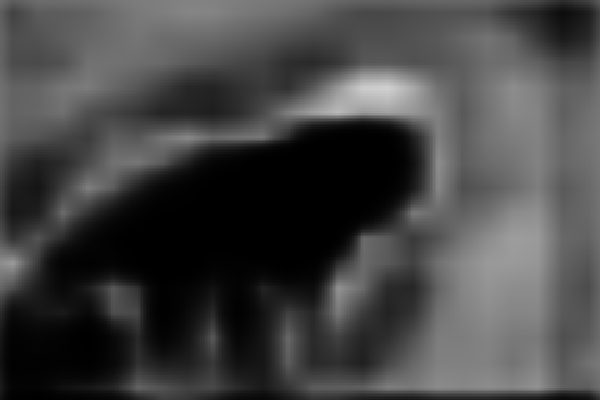}}\hskip.2em
    \subfloat[]{\includegraphics[width=0.16\textwidth]{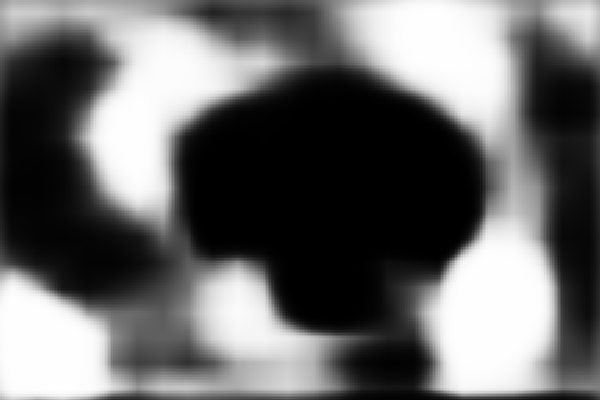}}\hskip.2em
    \subfloat[]{\includegraphics[width=0.16\textwidth]{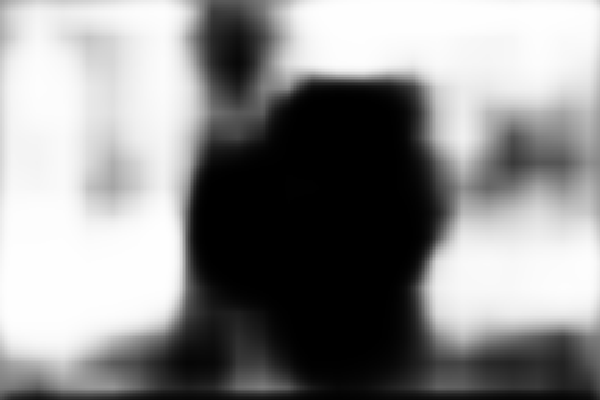}}\hskip.2em
    \subfloat[]{\includegraphics[width=0.16\textwidth]{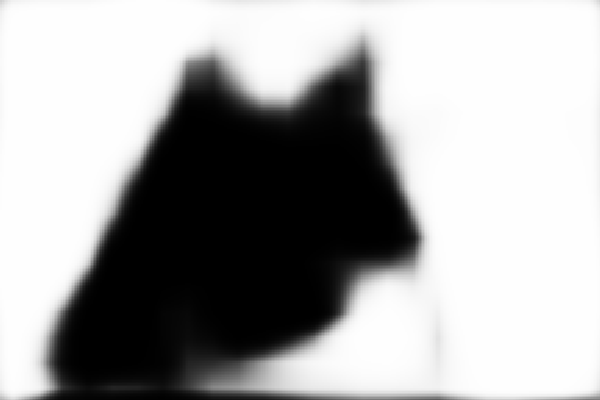}}\hskip.2em
    \subfloat[]{\includegraphics[width=0.16\textwidth]{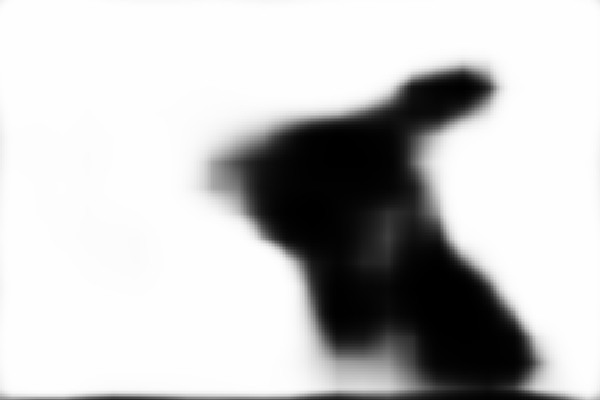}}\hskip.2em
    \subfloat[]{\includegraphics[width=0.16\textwidth]{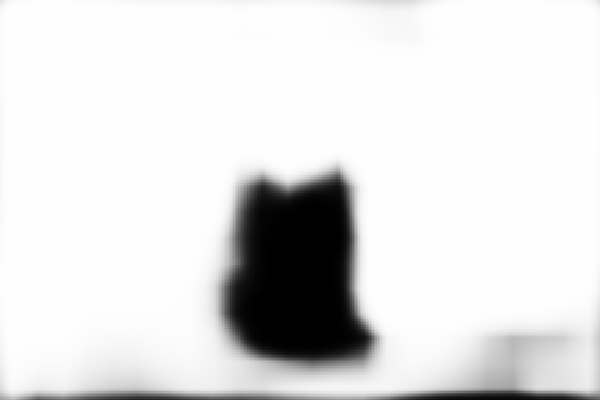}}

    \caption{The overall blur degree estimation based on our blur maps. The dog pictures in the first row are ranked from left to right according to the estimated blur degree $S$. (a) $S =
    0.286$. (b) $S=
    0.416$. (c) $S = 0.553$. (d) $S = 0.652$. (e) $S=
    0.767$. (f) $S=
    0.881$.}\label{fig:blurDegree}
\end{figure*}

\subsubsection{Blur Degree Estimation}
Our blur map can also serve as an estimation of the overall blur degree of an image. Since each entry in our blur map indicates the blur degree of the corresponding pixel in the image, we implement a straightforward blur degree measure $S$ of an image as the average value of the corresponding blur map. More sophisticated pooling strategies taking into account visual attention can also be incorporated to boost the performance.  Fig.~\ref{fig:blurDegree} shows a set of dog pictures ranked from left to right with increasing $S$, from which we can see that DBM robustly extracts blurred regions with high confidence and that the ranking results are in close agreement with human vision of blur perception.

%For partially blurred images, the relationship between the overall blur degree and their perceptual quality are non-trivial. Take the rightmost image in Fig~\ref{fig:blurDegree} as an example. Although the background that constitutes the main part of the image is blurred, its purpose is to give prominence to the foreground dog.  As a result, the blur does not degrade the quality of the image, instead, it makes the image more visually appearing, resulting in  nearly perfect quality. We conduct an experiment on the LIVE in the wild Challenge database~\cite{ghadiyaram2016massive} to explore the correlation between our blur degree measurement and the human opinions of image quality, and no strong correlation is observed as expected. Further investigations are needed to better incorporate the overall blur degree into the assessment of image quality, but it is out of scope of this work.

\subsubsection{Blur  Magnification} 
\begin{figure}
    \centering
    \captionsetup{justification=centering}

    \subfloat[]{\includegraphics[width=0.23\textwidth]{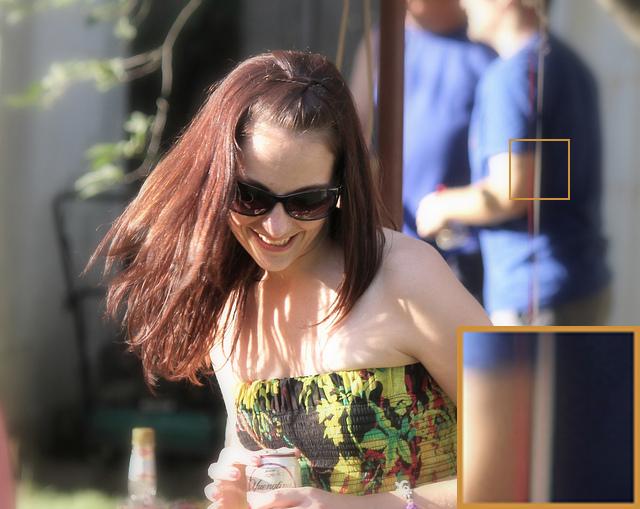}}\hskip1em
    \subfloat[]{\includegraphics[width=0.23\textwidth]{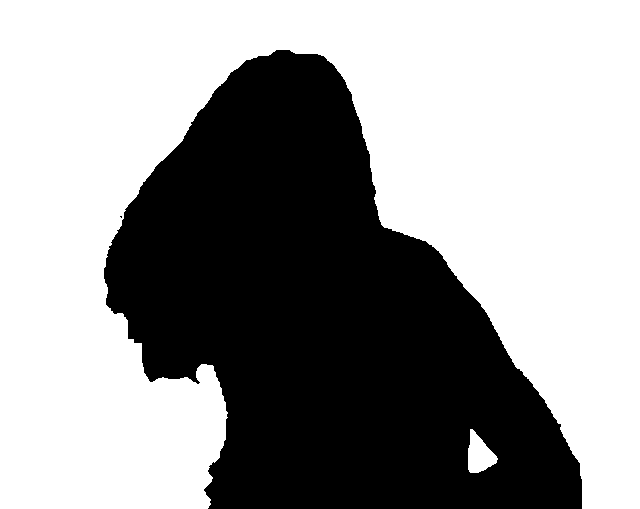}}\\
    \vspace{-9pt}
    \subfloat[]{\includegraphics[width=0.23\textwidth]{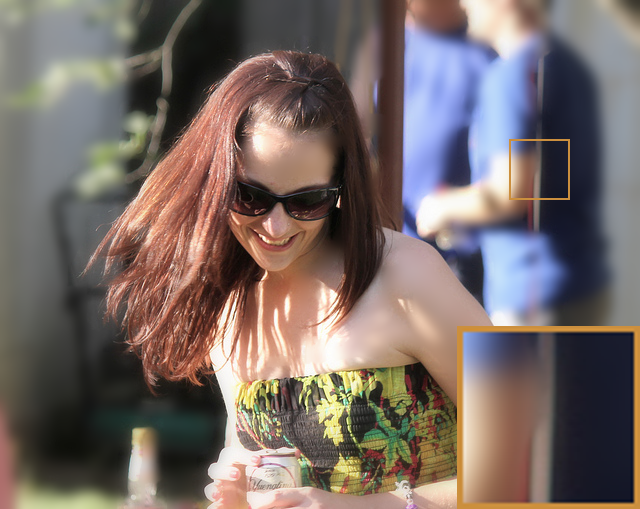}}\hskip1em 
    \subfloat[]{\includegraphics[width=0.23\textwidth]{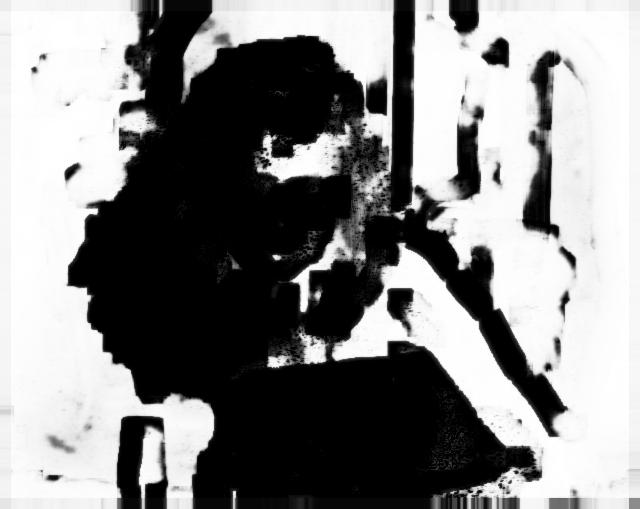}}\\
    \vspace{-9pt}
    \subfloat[]
    {\includegraphics[width=0.23\textwidth]{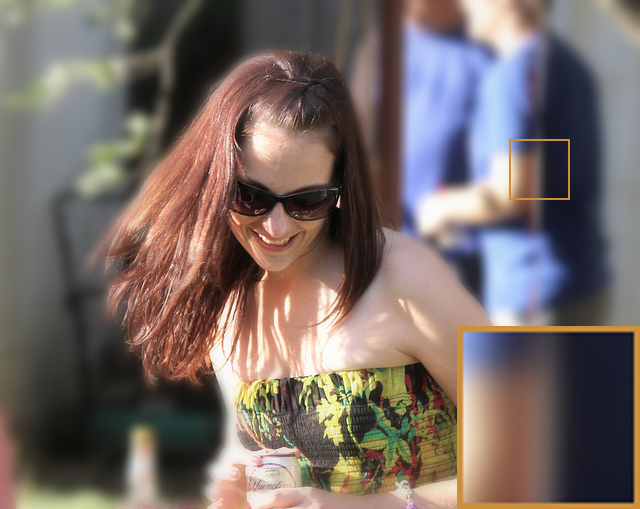}}\hskip1em
    \subfloat[]
    {\includegraphics[width=0.23\textwidth]{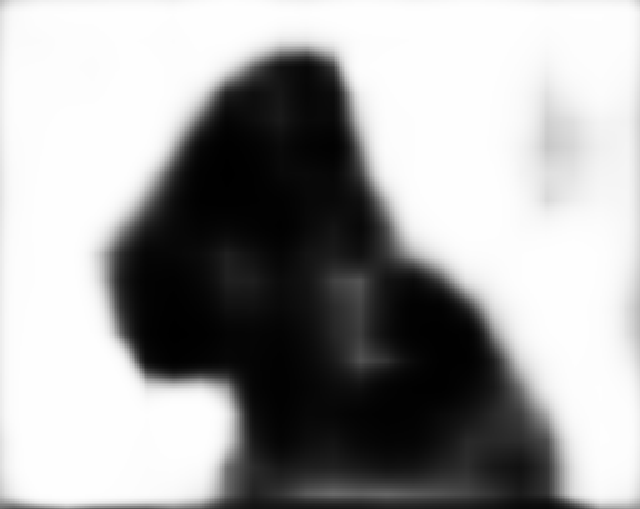}}
    \caption{The blur magnification results. (a) Test image from the blur detection benchmark~\cite{shi2014discriminative}. (b) Ground truth blur map. (c) Magnification by Shi14~\cite{shi2014discriminative}. (d) Blur map by Shi14~\cite{shi2014discriminative}. (e) Magnification by DBM. (f) Blur map by DBM. }\label{fig:blurMag}
\end{figure}

A shallow depth-of-field is often preferred in creative photography, such as portraits. However, current small cameras embedded in mobile devices limit the degree of defocus blur due to  small diameters of their lenses.  With extracted blurred regions, it is easy to drive a computational photography approach to increase defocus for blur magnification~\cite{bae2007defocus}. Here we implement a na{\"i}ve blur magnifier by convolving pixels with blur confidence greater than $0.1$ using a uniform Gaussian kernel. We compare DBM with Shi14~\cite{shi2014discriminative} in Fig.~\ref{fig:blurMag}. It is clear that DBM is barely affected by the structures with blurring and delivers a perceptually more  consistent result with smooth transitions from clear to blur regions.

\section{Conclusion and Discussion}\label{sec:conclusion}
In this paper, we explore visual blur mapping of natural images, emphasizing on the importance of high-level semantic information.  We opt for CNNs as a proper tool to explore high-level features, and develop the first end-to-end and image-to-image blur mapper based on an FCN. The proposed DBM significantly outperforms previous methods and successfully resolves challenging ambiguities such as differentiating flat and blurred regions. In the future, it remains to be seen how the low-level features and high-level semantics interplay with each other and how they can be used to predict visual blur perception.

%-------------------------------------------------------------------------

\begin{figure}
    \centering
    \captionsetup{justification=centering}
    \subfloat[]{\includegraphics[width=0.16\textwidth]{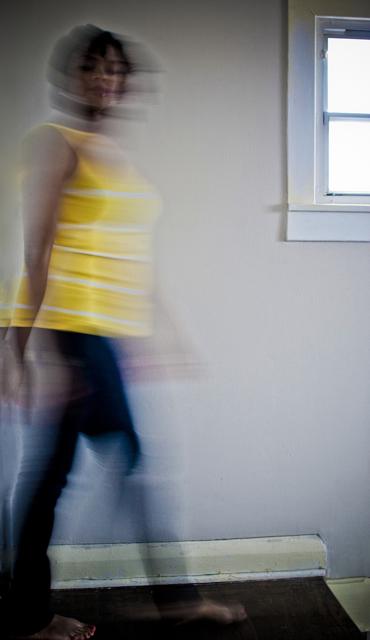}}\hskip.1em
    \subfloat[]{\includegraphics[width=0.16\textwidth]{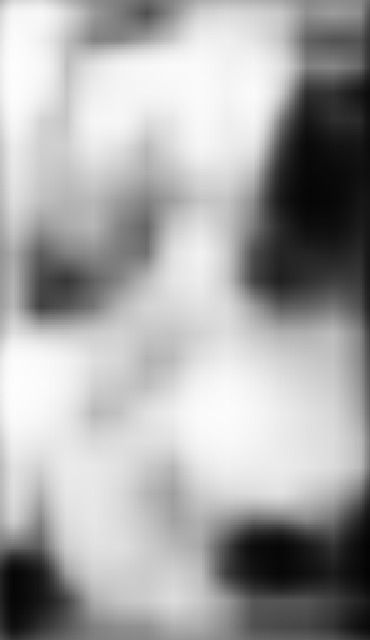}}\hskip.1em
    \subfloat[]{\includegraphics[width=0.16\textwidth]{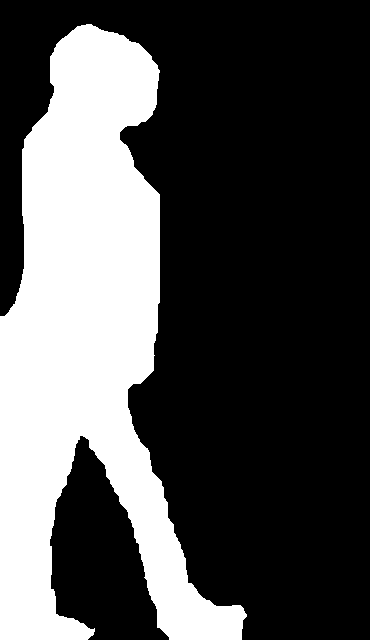}}
    \caption{Failure case of DBM. (a) Test image from the blur detection benchmark~\cite{shi2014discriminative}. (b) Blur map produced by DBM. (c) Ground truth.}\label{fig:fail}
\end{figure}

DBM fails occasionally in some cases. For example, if the motion-blurred subject happens to be surrounded by a large flat background, as shown in Fig.~\ref{fig:fail}, it is difficult to extract accurate and useful semantic information for local blur mapping. A potential solution is to retrain DBM on a larger database of more scene structure variations. Another limitation of DBM is that it generates blur maps with coarse boundaries. This may be improved by simultaneously learning a reconstruction network for boundary refinement~\cite{ghiasi2016laplacian}. These issues will be investigated in our future research.

\section*{Acknowledgements} The authors would like to thank Dr. Wangmeng Zuo and Dr. Dongwei Ren for deeply insightful comments on blur perception, Kai Zhang and Faqiang Wang for sharing their expertise on CNN, and Zhengfang Duanmu for helpful advices on debugging. We thank the NVIDIA Corporation for donating a GPU  for this research.

\ifCLASSOPTIONcaptionsoff
  \newpage
\fi



%\bibliographystyle{IEEEtran}

\bibliographystyle{IEEEtran}
\bibliography{egbib}

\begin{IEEEbiography}[{\includegraphics[width=1in,height=1.25in,clip,keepaspectratio]{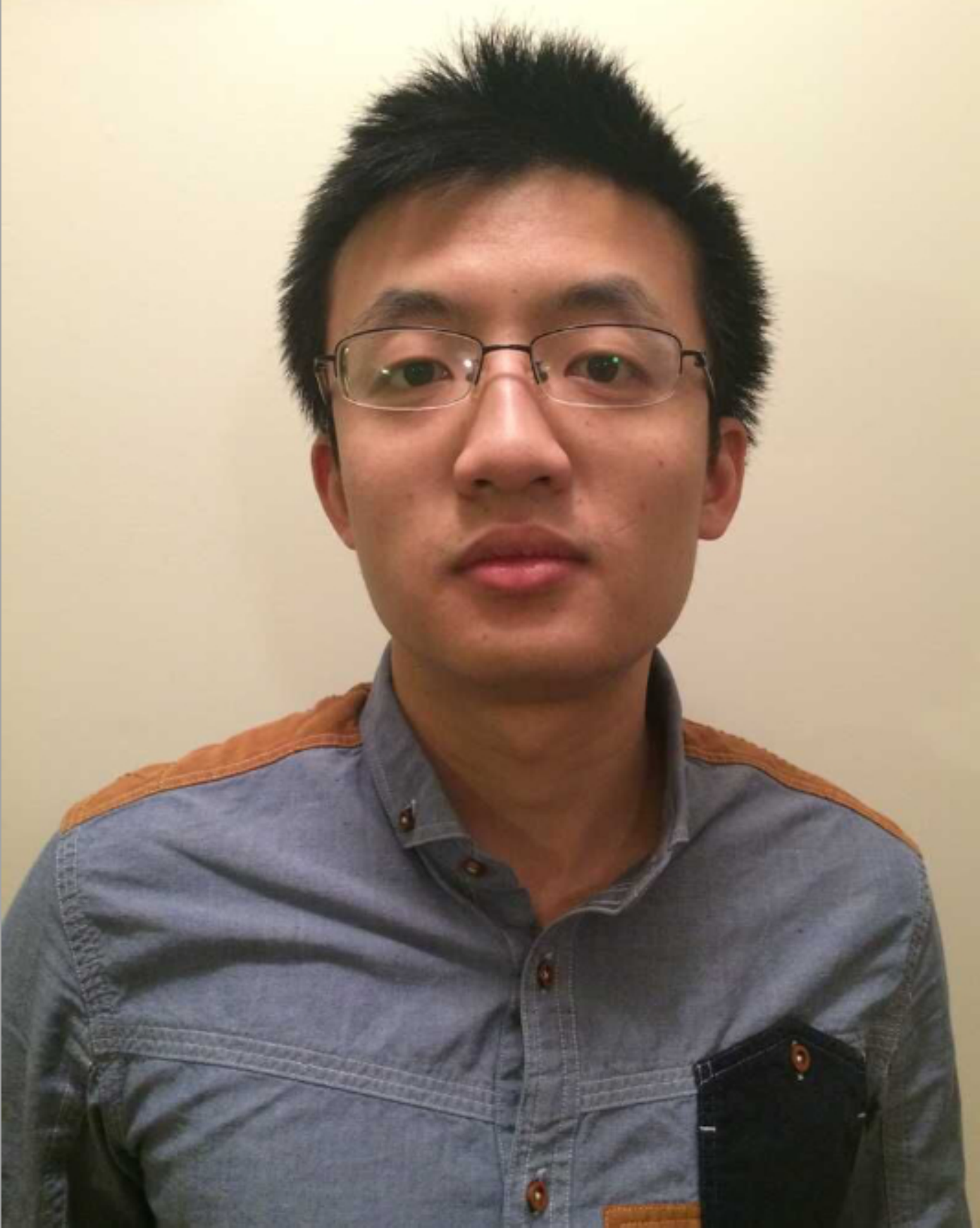}}]{Kede Ma}
(S'13-M'18) received the B.E. degree from the University of Science and Technology of China, Hefei, China, in 2012, and the M.S. and Ph.D. degrees in electrical and computer engineering from the University of Waterloo, Waterloo, ON, Canada, in 2014 and 2017, respectively. He is currently a Research Associate with Howard Hughes Medical Institute and  Laboratory for Computational Vision, New York University, New York, NY, USA. His research interests include perceptual image processing, computational vision, and computational photography. 
\end{IEEEbiography}

\begin{IEEEbiography}[{\includegraphics[width=1in,height=1.25in,clip,keepaspectratio]{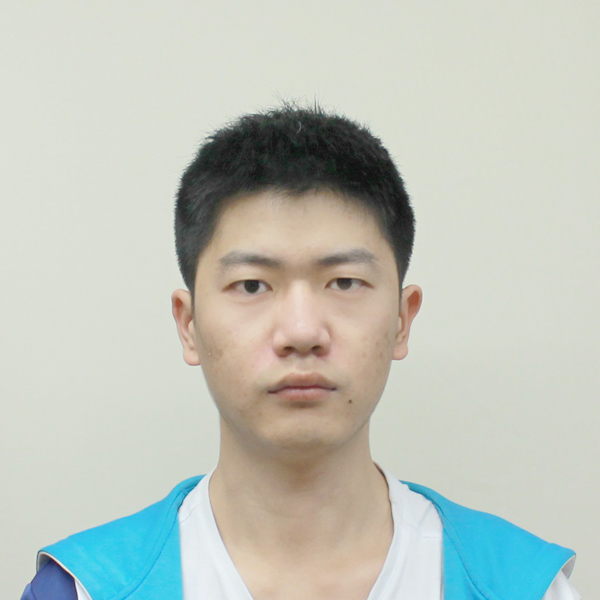}}]{Huan Fu}
received the B.E. degree from the University of Science and Technology of China, Hefei, China, in 2014. He is currently a Ph.D. student with the School of Information Technologies and the Faculty of Engineering and Information Technologies, and a member of the UBTECH Sydney AI Centre, The University of Sydney. His research interests include deep learning and computer vision.
\end{IEEEbiography}

% \begin{IEEEbiography}[{\includegraphics[width=1in,height=1.25in,clip,keepaspectratio]{bios/liu}}]{Tongliang Liu}
% received the B.Eng. degree in electronic engineering and information science from the University of Science and Technology of China, and the Ph.D. degree from the University of Technology Sydney. He is currently a Lecturer with
% the School of Information Technologies, Faculty of Engineering and Information Technologies, and a Core Member with the UBTech Sydney Artificial Intelligence Institute,
% The University of Sydney. His research interests include statistical learning theory, computer
% vision, and optimization. He has authored and co-authored over 40 research papers, including IEEE T-PAMI, T-NNLS, T-IP, ICML, CVPR, and KDD.
% \end{IEEEbiography}

\begin{IEEEbiography}[{\includegraphics[width=1in,height=1.25in,clip,keepaspectratio]{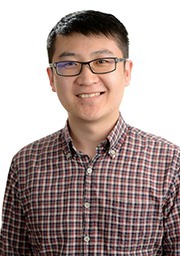}}]{Tongliang Liu} is
currently a Lecturer with the School of Information Technologies and the Faculty of Engineering and Information Technologies, and a core member in the UBTECH Sydney AI Centre, at The University of Sydney. He received the BEng degree in electronic engineering and information science from the University of Science and Technology of China, and the PhD degree from the University of Technology Sydney. His research interests include statistical learning theory, computer vision, and optimisation. He has authored and co-authored 40+ research papers including IEEE T-PAMI, T-NNLS, T-IP, ICML, CVPR, and KDD.
\end{IEEEbiography}

\begin{IEEEbiography}[{\includegraphics[width=1in,height=1.25in,clip,keepaspectratio]{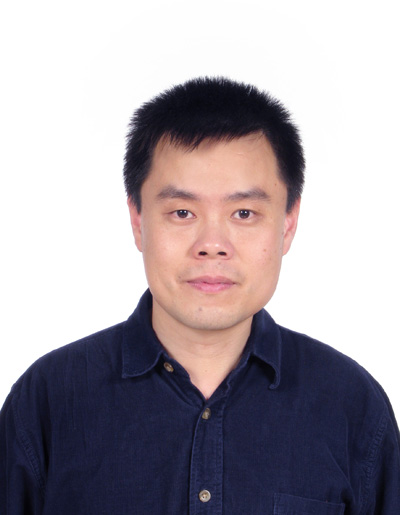}}]{Zhou Wang}
(S'99-M'02-SM'12-F'14) received the Ph.D. degree from The University of Texas at Austin, TX, USA, in 2001. He is currently a Professor and University Research Chair with the Department of Electrical and Computer Engineering, University of Waterloo, Waterloo, ON, Canada. His research interests include image and video processing and coding; visual quality assessment and optimization; computational vision and pattern analysis; multimedia communications; and biomedical signal processing. He has more than 200 publications in these fields with over 40,000 citations (Google Scholar).

Dr. Wang serves as a Senior Area Editor for the {\scshape IEEE Transactions on Image Processing} (2015-present), and an Associate Editor for the {\scshape IEEE Transactions on Circuits and Systems for Video Technology} (2016-present). Previously, he served as a member for the {\scshape IEEE Multimedia Signal Processing Technical Committee} (2013-2015), an Associate Editor for the {\em IEEE Transactions on Image Processing} (2009-2014), {\scshape Pattern Recognition} (2006-present) and {\scshape IEEE Signal Processing Letters} (2006-2010), and a Guest Editor of {\scshape IEEE Journal of Selected Topics in Signal Processing} (2013-2014 and 2007-2009). He is a Fellow of Canadian Academy of Engineering, and a recipient of the 2017 Faculty of Engineering Research Excellence Award at University of Waterloo, the 2016 IEEE Signal Processing Society Sustained Impact Paper Award, the 2015 Primetime Engineering Emmy Award, the 2014 NSERC E.W.R. Steacie Memorial Fellowship Award, the 2013 IEEE Signal Processing Magazine Best Paper Award, and the 2009 IEEE Signal Processing Society Best Paper Award.
\end{IEEEbiography}

\begin{IEEEbiography}[{\includegraphics[width=1in,height=1.25in,clip,keepaspectratio]{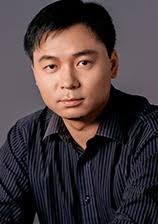}}]{Dacheng Tao}
(F'15) is Professor of Computer Science and ARC Laureate Fellow in the School of Information Technologies and the Faculty of Engineering and Information Technologies, and the Inaugural Director of the UBTECH Sydney Artificial Intelligence Centre, at the University of Sydney. He mainly applies statistics and mathematics to Artificial Intelligence and Data Science. His research interests spread across computer vision, data science, image processing, machine learning, and video surveillance. His research results have expounded in one monograph and 500+ publications at prestigious journals and prominent conferences, such as IEEE T-PAMI, T-NNLS, T-IP, JMLR, IJCV, NIPS, ICML, CVPR, ICCV, ECCV, ICDM; and ACM SIGKDD, with several best paper awards, such as the best theory/algorithm paper runner up award in IEEE ICDM’07, the best student paper award in IEEE ICDM’13, the distinguished student paper award in the 2017 IJCAI, the 2014 ICDM 10-year highest-impact paper award, and the 2017 IEEE Signal Processing Society Best Paper Award. He received the 2015 Australian Scopus-Eureka Prize, the 2015 ACS Gold Disruptor Award and the 2015 UTS Vice-Chancellor’s Medal for Exceptional Research. He is a Fellow of the Australian Academy of Science, AAAS, IEEE, IAPR, OSA and SPIE.
\end{IEEEbiography}

\end{document}